\documentclass[11pt]{article}
\usepackage{cite}
\usepackage{amsmath,amssymb,amsfonts}
\usepackage{algorithmic}
\usepackage{graphicx}
\usepackage{textcomp}
\usepackage{xcolor}
\usepackage{subfigure}
\usepackage{algorithm}
\usepackage{booktabs}
\usepackage[backref]{hyperref}
\usepackage[margin=1in]{geometry}
\def\BibTeX{{\rm B\kern-.05em{\sc i\kern-.025em b}\kern-.08em
    T\kern-.1667em\lower.7ex\hbox{E}\kern-.125emX}}

\title{A Transformer-based Diffusion Probabilistic Model for Heart Rate and Blood Pressure Forecasting in Intensive Care Unit}

\author{Ping Chang$^1$, \and Huayu Li$^1$, \and Stuart F. Quan$^{2,3}$, \and Shuyang Lu $^{4,5}$, \and Shu-Fen Wung $^{6,7}$, \and Janet Roveda$^{1,6,8}$, \and Ao Li$^{1,6}$}
\date{}

\begin{document}
\maketitle
\noindent
$^1$Department of Electrical \& Computer Engineering, The University of Arizona, Tucson, AZ, USA\\
$^2$Division of Sleep and Circadian Disorders, Departments of Medicine and Neurology, Brigham and Women’s Hospital, Harvard Medical School, Boston, MA, USA\\
$^3$Asthma and Airway Disease Research Center, College of Medicine, The University of Arizona, Tucson, AZ, USA\\
$^4$Department of Cardiovascular Surgery, Zhongshan Hospital, Fudan University, Shanghai, P.R. China \\
$^5$The Shanghai Institute of Cardiovascular Diseases, Shanghai, P.R. China\\
$^6$Bio5 Institute, The University of Arizona, Tucson, AZ, USA\\
$^7$College of Nursing, The University of Arizona, Tucson, AZ, USA\\
$^8$Department of Biomedical Engineering, The University of Arizona, Tucson, AZ, USA\\[30pt]
Corresponding author:\\
Ao Li, Ph.D.\\
Department of Electrical and Computer Engineering\\
The University of Arizona\\
1230 E Speedway Blvd\\
Tucson, AZ, 85719\\
Email: aoli1@arizona.edu

\clearpage
\begin{abstract}
\noindent \textbf{Background and Objective:} 
Vital sign monitoring in the Intensive Care Unit (ICU) is crucial for enabling prompt interventions for patients. This underscores the need for an accurate predictive system. Therefore, this study proposes a novel deep learning approach for forecasting Heart Rate (HR), Systolic Blood Pressure (SBP), and Diastolic Blood Pressure (DBP) in the ICU.

\noindent \textbf{Methods:} We extracted $24,886$ ICU stays from the MIMIC-III database which contains data from over $46$ thousand patients, to train and test the model. The model proposed in this study, Transformer-based Diffusion Probabilistic Model for Sparse Time Series Forecasting (TDSTF), merges Transformer and diffusion models to forecast vital signs. The TDSTF model showed state-of-the-art performance in predicting vital signs in the ICU, outperforming other models' ability to predict distributions of vital signs and being more computationally efficient. The code is available at https://github.com/PingChang818/TDSTF.

\noindent \textbf{Results:} The results of the study showed that TDSTF achieved a Standardized Average Continuous Ranked Probability Score (SACRPS) of $0.4438$ and a Mean Squared Error (MSE) of $0.4168$, an improvement of $18.9\%$ and $34.3\%$ over the best baseline model, respectively. The inference speed of TDSTF is more than $17$ times faster than the best baseline model.

\noindent \textbf{Conclusion:} TDSTF is an effective and efficient solution for forecasting vital signs in the ICU, and it shows a significant improvement compared to other models in the field.

\end{abstract}

\clearpage

\section{Introduction}
Vital signs are crucial in monitoring patients' health and body functions in the ICU. Continuous monitoring systems alert caregivers of potential adverse events \cite{kenzaka2012importance, yoon2019predicting}. A predictive warning system for vital signs can save valuable time by enabling prompt interventions \textcolor{black}{\cite{subbe2001validation, sessler2019beyond}}. However, the implementation of such a system faces several challenges. ICUs are complex environments where patients have multiple underlying conditions, treatments, and interventions that can affect their vital signs, making it difficult to develop algorithms that accurately predict them \textcolor{black}{\cite{doig2011informing, collins2012search}}. Vital sign prediction requires large amounts of data to develop accurate algorithms \textcolor{black}{\cite{kristinsson2022prediction}}, but in many ICUs, the availability and quality of electronic health record data can be limited \textcolor{black}{\cite{ghassemi2015multivariate, tipirneni2022self}}. This makes it challenging to use classical statistical methods, which often rely on manual feature engineering and cannot capture complex patterns \textcolor{black}{\cite{unanue2017recurrent, ij2018statistics}}. As a result, few attempts have been made to predict vital signs in the ICU using these methods.

With the development of deep learning, applying this new approach to predicting ICU vital signs is possible. As is well known, deep learning has revolutionized the field of time series forecasting in recent years, with its high representational power enabling successful predictions of vital signs in various studies. For instance, in Generative Boosting \cite{liu2019early}, the Long Short-Term Memory (LSTM) network is used to create a generative model, effectively reducing error propagation and improving HR prediction performance. In a comparison of Recurrent and Convolutional Neural Network (RNN and CNN)  \cite{masum2019investigation}, using different horizon strategies on the MIMIC-II dataset, the bidirectional LSTM (Bi-LSTM) with the DIRMO strategy delivered the best predictions of HR and blood pressure. The TOP-Net model \cite{liu2021top} predicts tachycardia onset using Bi-LSTM, with a prediction horizon of up to $6$ hours. It was trained on the data of less than $6$ thousand patients from the MIMIC-III database. The Temporal Fusion Transformer \cite{phetrittikun2021temporal} predicts vital sign quantiles based on an attention mechanism, capturing anomaly temporal patterns and optimizing input time windows by calculating temporal importance.

\textcolor{black}{Although these methods have shown promising results in vital sign forecasting, they still face several limitations when it comes to practical application in the ICU setting. First, these models require continuous monitoring of vital signs. However, given the complex and unpredictable conditions in the ICU, monitoring often becomes intermittent and sporadic.}  Second, these models only consider vital signs, ignoring the interrelated events\textemdash interventions and conditions surrounding a patient in the ICU\textemdash that could improve forecasting accuracy. For example, the existing models should have addressed active interventions such as medications, potential medical procedures, and their impact on patient vital signs. Third, the datasets used to evaluate these models often have a limited number of subjects, leading to a lack of generalizability and potential bias, which is a major concern in critical ICU scenarios.

\textcolor{black}{We aim to develop an effective and efficient deep-learning approach to forecast
Heart Rate (HR), Systolic Blood Pressure (SBP), and Diastolic Blood Pressure (DBP) in the ICU setting.} The use of diffusion probabilistic models (diffusion models for short) \cite{rasul2021autoregressive, tashiro2021csdi} in time series analysis has been gaining popularity due to their balance between flexibility and tractability \cite{sohl2015deep}. These models have achieved state-of-the-art performance in time series forecasting. Our study aims to investigate the effectiveness of diffusion models in handling sparse time series data and making fast and accurate predictions of vital signs in the ICU setting. The triplet form of the diffusion model enhances its ability to process sparse data, and using a Transformer-based backbone leads to improved performance compared to baseline models. The ultimate goal is to promptly provide ICU caregivers with critical information by considering all recorded events, not just vital signs. The novel contributions of this paper include: (1) An examination of the ability of the diffusion model to extract temporal dependencies from sparse time series data. (2) Use the triplet form to enhance the efficiency of the diffusion model when processing sparse data. (3) Fast and accurate forecasting of vital signs in the ICU setting. (4) Integration of all recorded events in the ICU setting for vital sign forecasting without being limited by the screened data. (5) Comparison of the proposed model with baseline models, showing improved performance using the Transformer as the backbone.
\section{Related Works}
\subsection{Probabilistic Time Series Forecasting}
In many real-world scenarios, the future is uncertain, and making a single best estimate of the outcome is impossible. To account for this uncertainty, probabilistic forecasting provides a range of possible outcomes and probabilities of each. This is particularly significant in the ICU setting, where caregivers must understand the risks associated with different decisions. Several probabilistic time series forecasting models have been proposed in recent years and achieved state-of-the-art performance. MQ-RNN \cite{wen2017multi} uses RNN to generate hidden states of the input time series, which are then transformed into contextual information by a global Multilayer Perceptron Network (MLP). The local MLP uses contextual information and covariates to generate quantile predictions. In DeepAR \cite{salinas2020deepar}, hidden states are obtained through RNN, which are then input into linear layers with activation to generate the mean and variance of the assumed likelihood model that samples predictions. DeepFactor \cite{wang2019deep} assumes that time series are exchangeable and decomposes the joint distribution into global and local time series. RNN is subsequently employed to capture global non-linear patterns. Simultaneously, an assumed observation model is applied to discern local uncertainties conditioned on the global effects. The outputs of these two functions are used to generate the forecasting distribution. \textcolor{black}{EnCQR \cite{jensen2022ensemble} is a homogeneous ensemble approach. Member learners within this framework can be constructed utilizing various machine learning algorithms. EnCQR combines conformal prediction and quantile regression methodologies to construct prediction intervals devoid of reliance on specific distributional assumptions.}

\subsection{Diffusion Model}
The Diffusion model is a powerful generative model that learns the underlying distribution of data by transforming data samples into Gaussian noise, and has achieved state-of-the-art performance in various applications \textcolor{black}{\cite{yang2023diffusion}}. Initially, it garnered significant attention due to its superior performance in image synthesis compared to the Generative Adversarial Network (GAN) \cite{ho2020denoising, song2020denoising, austin2021structured}. In recent years, its potential has expanded to domains such as protein sequence analysis \cite{anand2022protein}, threat detection \cite{blau2022threat}, audio synthesis \cite{oord2016wavenet}, and probabilistic time series forecasting \cite{tashiro2021csdi, rasul2021autoregressive}. CSDI takes as input a matrix filled with both historical data and target and a mask matrix indicating missing values \cite{tashiro2021csdi}. Its backbone is based on DiffWave \cite{kong2020diffwave}, which enables correlation across all features and time points. The results from CSDI has shown that the diffusion model can be optimized by selecting the appropriate backbone for specific tasks.

\subsection{Transformer}
The Transformer uses an encoder-decoder architecture \cite{vaswani2017attention}, widely applied in Natural Language Processing (NLP) tasks. It is known for its ability to attend to specific parts of the input sequence rather than considering the entire sequence equally. This is achieved through the use of an attention mechanism, which assigns a weight to each element of the input sequence, indicating the amount of attention the model should allocate to each element when making predictions. A notable extension of the Transformer is GPT-$3$ \cite{brown2020language}, which has demonstrated strong semantic representation capabilities. ICU data share many characteristics with NLP data, given the vast capacity of dictionaries. While multiple aspects of a patient's condition can be monitored in the ICU, only a limited number of them are recorded at any given time. Additionally, the recording intervals may be irregular, presenting data processing challenges. Furthermore, different patients may have different items recorded, and all possible items must be considered in the analysis. All of these factors contribute to the extreme sparsity of ICU data \textcolor{black}{\cite{ghassemi2015multivariate, tipirneni2022self}}.
\section{Methods}
\subsection{TDSTF}
Generative models aim to learn the underlying distribution of an observation dataset, but the main challenge lies in marginalizing out the latent variables to calculate the normalizing constant for a valid distribution, which is intractable \cite{zhang2018advances}. Variational inference is a commonly used solution, transforming the distribution calculation into an optimization problem. The diffusion model applies variational inference to approximate a data distribution. It is based on the idea that a continuous Gaussian diffusion process can be reversed with the same functional form as the forward process \cite{feller2015theory}. After learning the reverse process, the input pure noise will converge to data points sampled from the modeled distribution. This can be approximated with a discrete Gaussian diffusion process, given a large enough number of $T$ diffusion steps.

\begin{figure}
\centering
\includegraphics[width=.9\textwidth]{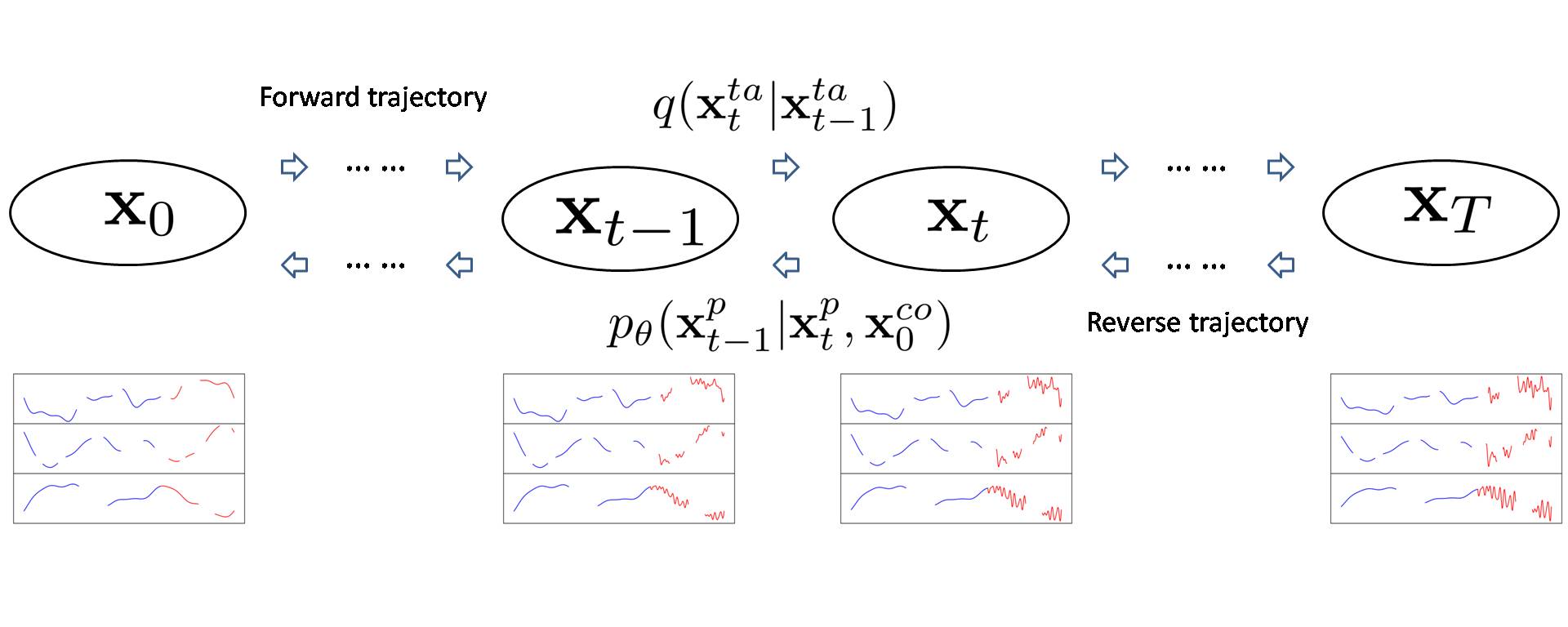}
\caption{Diagram of the diffusion processes in our forecasting model. The blue curves indicate the history events as the conditional data $\mathbf{x}_0^{co}$. The red curves symbolize the noisy target $\mathbf{x}_t^{ta}$ at time step $t$ in the forward trajectory or the intermediate result $\mathbf{x}_t^p$ during prediction. The target is represented by $\mathbf{x}_0^{ta}$. The gaps among the curves symbolize the missing values in the sparse data. The forward trajectory $q$ adds noise of increasing levels to $\mathbf{x}_0^{ta}$. The reverse trajectory $p_\theta$ then removes the noise from the pure noise $\mathbf{x}_T^{p}$ to generate samples.}
\label{fig:diffusion_process}
\end{figure}

We propose a Transformer-based Diffusion Probabilistic Model for Sparse Time Series Forecasting (TDSTF). The diffusion processes in terms of time series forecasting are manifested in Figure \ref{fig:diffusion_process}, divided into forward and reverse trajectories. Conditioned on the history observation $\mathbf{x}_0^{co}$, The purpose of our method is to learn $p_ \theta (\mathbf{x}_0^p|\mathbf{x}_0^{co})$ parameterized by $\theta$ that approximates $q(\mathbf{x}_0^{ta}|\mathbf{x}_0^{co})$, so that $\mathbf{x}_0^p$ predicts the target $\mathbf{x}_0^{ta}\sim q(\mathbf{x}_0^{ta}|\mathbf{x}_0^{co})$. During forward trajectory, a small amount of noise is added to the \textcolor{black}{ground truth} data $\mathbf{x}_0^{ta}$ at each step $t$ \textcolor{black}{to obtain the noisy target $\mathbf{x}_t^{ta}$. At the final step $T$, $\mathbf{x}_T^{ta}\sim N(\mathbf{0},\mathbf{I})$}. The noise amount at each step is determined by a variance schedule $\beta _1,...,\beta _T$. The forward process is expressed as a Markov chain:
\begin{equation}
\label{eq:forward}
\begin{aligned}
&q(\mathbf{x}_{1:T}^{ta}|\mathbf{x}_0^{ta}):=\prod _{t=1}^{T}q(\mathbf{x}_t^{ta}|\mathbf{x}_{t-1}^{ta})\\
&q(\mathbf{x}_t^{ta}|\mathbf{x}_{t-1}^{ta}):=N(\mathbf{x}_t^{ta};\sqrt {1-\beta _t} \mathbf{x}_{t-1}^{ta},\beta _t\mathbf{I})
\end{aligned}
\end{equation}
\noindent where $N$ stands for Gaussian distribution parameterized by $\sqrt {1-\beta _t} \mathbf{x}_{t-1}^{ta}$ as its mean and $\beta _t\mathbf{I}$ as its variance. Table \ref{tab:notation} lists all notations used in our method for convenience. The bold font denotes a scalar vector. For instance, $\mathbf{x}=(x_1,x_2,...,x_n)$ where $n$ is the dimension of the vector.

\begin{table}
\caption{Notations used in the proposed model.}
\label{tab:notation}
\begin{center}
\begin{tabular}{cl}
\toprule
Notation&Description\\
\midrule
\textcolor{black}{$\mathbf{x}_0^{ta}$}&\textcolor{black}{Ground truth of target without noise}\\
$\mathbf{x}_t^{ta}$&Noisy target during forward trajectory at step $t$\\
$\mathbf{x}_t^p$&Intermediate result during inference at step $t$\\
$\mathbf{x}_0^{co}$&Conditional data\\
$\epsilon _\theta$&Backbone network of the diffusion model parameterized by $\theta$\\
$q$&Distribution of noisy target\\
$p_\theta$&Distribution of output from the diffusion model\\
$\beta _t$&Diffusion schedule at step $t$\\
$T$&Number of diffusion steps\\
\bottomrule
\end{tabular}
\end{center}
\end{table}

The reverse process begins with a Gaussian noise sample $\mathbf{x}_T^p\sim N(\mathbf{0},\mathbf{I})$. At each time step, the process progressively denoises $\mathbf{x}_t^p$ until $\mathbf{x}_0^p$ is obtained. The final sampled $\mathbf{x}_0^p$ is drawn from a distribution designed to resemble the training data distribution. This process can also be represented as a Markov chain:
\begin{equation}
\label{eq:reverse}
\begin{aligned}
&p_\theta (\mathbf{x}_{0:T}^p|\mathbf{x}_0^{co}):=p(\mathbf{x}_T^p)\prod _{t=1}^{T}p_\theta (\mathbf{x}_{t-1}^p|\mathbf{x}_t^p,\mathbf{x}_0^{co})\\
&p_\theta (\mathbf{x}_{t-1}^p|\mathbf{x}_t^p,\mathbf{x}_0^{co}):=N(\mathbf{x}_{t-1}^p;\mu _\theta (\mathbf{x}_t^p,t|\mathbf{x}_0^{co}),\Sigma _\theta)
\end{aligned}
\end{equation}
\noindent where $\mu_\theta$ and $\Sigma_\theta$ are learnable functions that generate the mean and variance of the modeled distribution. Equation \ref{eq:forward} implies that $\mathbf{x}_t^{ta}=\sqrt{\hat{\alpha}_t}\mathbf{x}_{t-1}^{ta} + \sqrt{1-\hat{\alpha}_t}\epsilon = \sqrt{\alpha_t}\mathbf{x}_0^{ta} + \sqrt{1-\alpha_t}\epsilon$, where $\epsilon\sim N(\mathbf{0},\mathbf{I})$. This means that we can sample at any time step $t$ during the forward process, based only on $\mathbf{x}_0^{ta}$. As a result, it is advantageous to construct $\mu _\theta$ and $\Sigma _\theta$ as follows:
\begin{align}
&\mu _\theta (\mathbf{x}_t^p,t|\mathbf{x}_0^{co})=\frac{1}{\sqrt{\hat{\alpha}_t}}(\mathbf{x}_t^p-\frac{\beta _t}{\sqrt{1-\alpha _t}}\epsilon _\theta (\mathbf{x}_t^p,t|\mathbf{x}_0^{co}))\\
&\Sigma _\theta(t)=\sigma _t^2=\frac{1-\alpha _{t-1}}{1-\alpha _t}\beta _t\; \; (t>1)
\end{align}
\noindent where $\hat{\alpha}_t=1-\beta _t$ and $\alpha _t=\prod _{i=1}^{t}\hat{\alpha}_i$, in order for $p_\theta (\mathbf{x}_{t-1}|\mathbf{x}_t)$ to be close to $q(\mathbf{x}_{t-1}|\mathbf{x}_t,\mathbf{x}_0)$ as stated in \cite{luo2022understanding}.

\subsection{\textcolor{black}{Model Structure}}
\begin{figure}
\centering
\includegraphics[width=.9\textwidth]{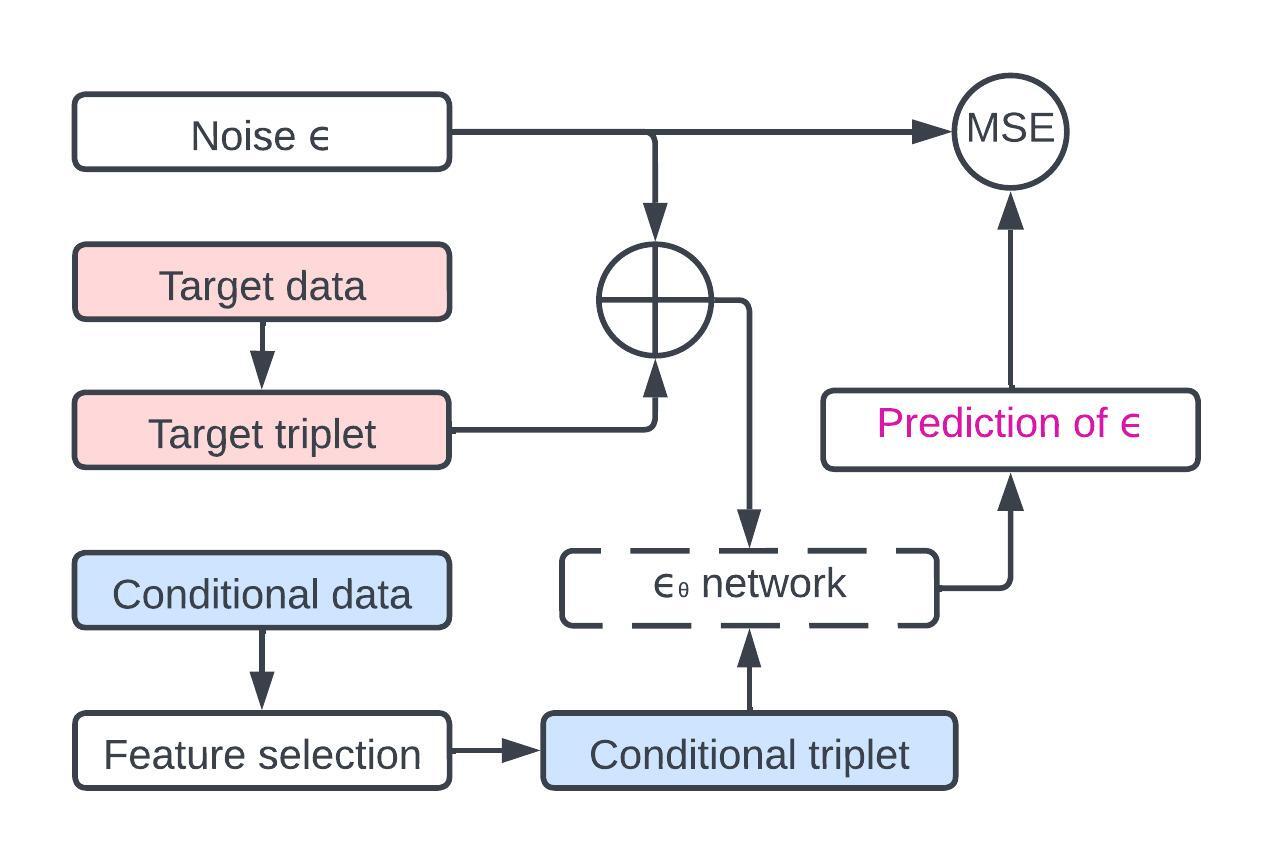}
\caption{The training procedure of our forecasting model \textcolor{black}{minimizes} the Mean Squared Error (MSE) between the noise prediction $\epsilon _\theta(x_t^{ta},t|x_0^{co})$ and $\epsilon$ as the loss function. The implementation of $\epsilon _\theta$ in the dashed box will be expanded and explained in detail later.}
\label{fig:training_procedure}
\end{figure}

The objective of the model training is to maximize the Evidence Lower Bound of $p_\theta (\mathbf{x}_0^p|\mathbf{x}_0^{co})$ \cite{ho2020denoising}. It can be expressed as the following equation:
\begin{equation}
\mathop {\min}_\theta E_t\Vert \epsilon -\epsilon _\theta (\mathbf{x}_t^{ta},t|\mathbf{x}_0^{co})\Vert _2 ^2
\end{equation}
\noindent where $\epsilon _\theta (\mathbf{x}_t^{ta},t|\mathbf{x}_0^{co})$ is a function that can be learned to predict $\epsilon$ for each step of the denoising process. Figure \ref{fig:training_procedure} illustrates the entire training procedure. Before being input into $\epsilon _\theta$, $\mathbf{x}_0^{ta}$ and $\mathbf{x}_0^{co}$ are transformed into triplets.

\begin{figure}
\centering
\includegraphics[width=.9\textwidth]{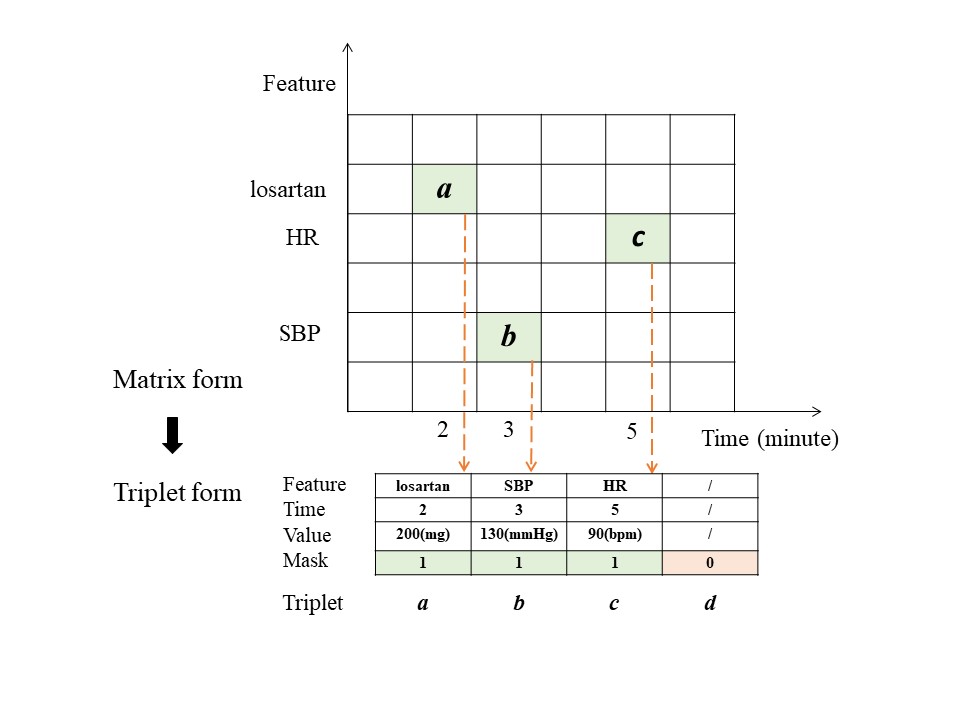}
\caption{An illustration of converting a sparse matrix to triplet representation. In this example, a patient received $200$ milligram (mg) of losartan at the second minute. A Heart Rate (HR) of $90$ beats per minute (bpm) and a Systolic Blood Pressure (SBP) of $130$ millimeters of mercury (mmHg) are recorded in the third and fifth minute, respectively. Assuming all other data points in the matrix are missing, and the $3$ valid event records $\textbf{\textit{a}}$, $\textbf{\textit{b}}$, and $\textbf{\textit{c}}$ line up in an array of triplets. \textcolor{black}{The size of the input triplet arrays is preset, and it can be larger than the number of valid triplets. The mask value of $0$ in $\textbf{\textit{d}}$ signifies the invalidity of this triplet, meaning the invalidity of its other $3$ elements.}}
\label{fig:triplet}
\end{figure}

Traditional methods, such as aggregation and imputation, are ineffective when dealing with extremely sparse data. Aggregation results in poor resolution and loss of temporal information, while imputation introduces excessive noise. To overcome these issues, the triplet form compactly stores sparse data. Each triplet contains a feature, time, value, and a mask bit indicating the presence or absence of data, represented by $1$ or $0$, respectively. The absolute time of the valid data points from the raw data is transformed into a relative time range. Figure \ref{fig:triplet} gives an illustration of converting a sparse matrix to a triplet representation. If the number of conditional triplets exceeds the input size (preset as a hyperparameter according to data preprocessing), the feature selection module prioritizes data points of the same features as the target, most correlated to the target data \cite{liu2021top, phetrittikun2021temporal}, and then fills in the remaining input triplets randomly. The Mean Squared Error (MSE) between the predicted noise and the Gaussian noise $\epsilon$ is used as the loss function.

\begin{figure}
\centering
\includegraphics[width=1\textwidth]{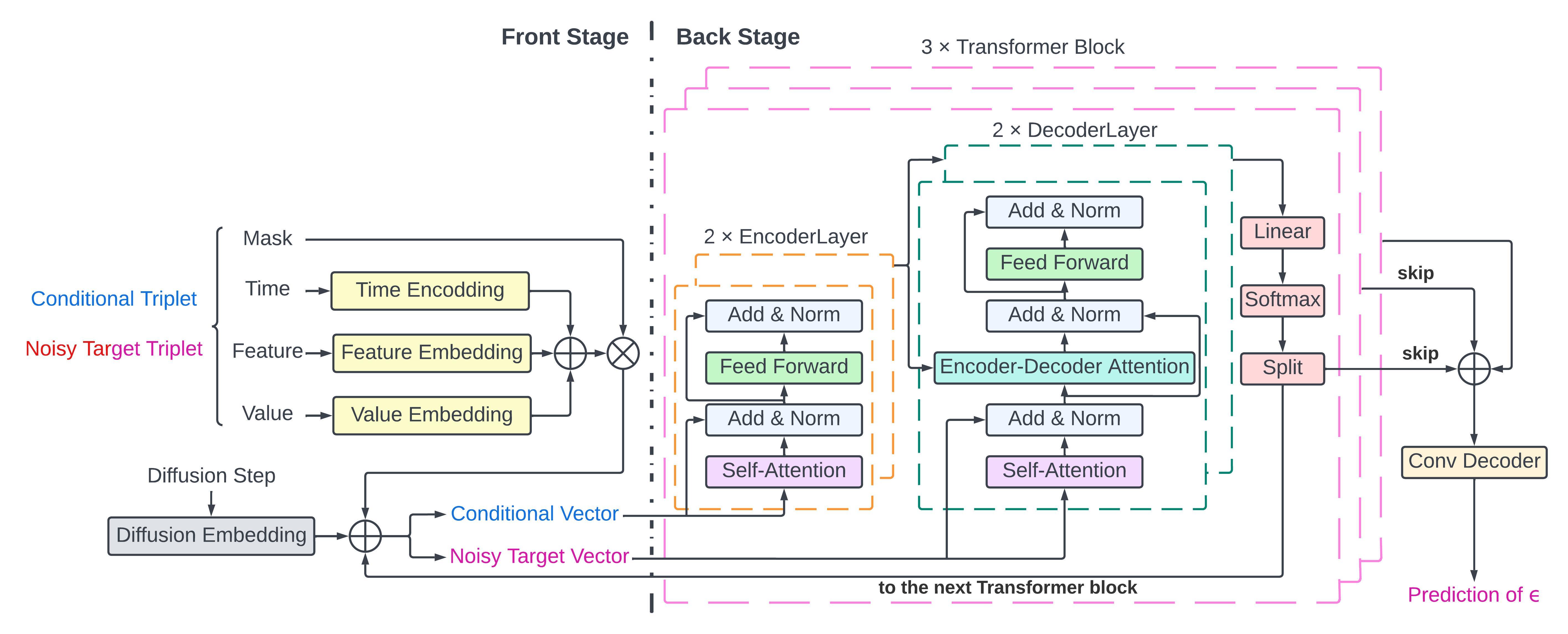}
\caption{The architecture of the $\epsilon_\theta$ network, designed for predicting $\epsilon$ at the current step.}
\label{fig:backbone_architecture}
\end{figure}

We construct $\epsilon_\theta$ using a deep neural network that is divided into two stages: the front stage and the back stage. This architecture is depicted in Figure \ref{fig:backbone_architecture}. The front stage maps the data into higher-dimensional spaces. A triplet's feature, value, and time components are transformed into vectors. The embedding module maps the triplet features into vectors, the linear layer projects the triplet values into vectors, and a group of sinusoidal functions transforms the triplet times into vectors. To avoid disturbing the missingness representations, the representations of the feature and time also incorporate information about the missingness, and the values of triplets with masks of $0$ are set to $0$ (the mean value for all features after \textcolor{black}{standardization}). A random diffusion step is applied in each iteration to generate noisy target values. The diffusion step is represented as a vector obtained from a lookup table and projected through linear layers. The results of all these vectors are fed into the back stage.

\textcolor{black}{Considering the trait similarity between the NLP data and the triplet data within our context, both originating from expansive and diverse spaces, we employ Transformer to build the back stage. This decision aligns with the prevailing discussions advocating for the attention mechanism to its efficacy in handling complex data representations \cite{brown2020language, devlin2018bert, vaswani2017attention}.} The conditional triplet is transformed by the encoders into Query ($Q$), Key ($K$), and Value ($V$) vectors. The self-attention layer calculates the correlation strength between one triplet and every other triplet through the following equation:
\begin{equation}
Attention=softmax(\frac{QK^T}{\sqrt{d_k}})V
\end{equation}
\noindent where $d_k$ is the dimension $Q$ and $K$, and the scaling factor $\sqrt{d_k}$ enhances the model's robustness. $K$ and $V$ output from the top encoder layer are input into the decoder's encoder-decoder attention layers, together with the target $Q$, to obtain the cross attention. The multi-headed attention mechanism also enhances the model's representational power by expanding latent subspaces with multiple independent sets of $Q$, $K$, and $V$. \textcolor{black}{The $\epsilon_\theta$ architecture is characterized by its use of $3$ Transformer blocks, each consisting of $2$ stacked encoder-decoder Transformer that sequentially process the input data. This encoder–decoder pattern can refine the output of the first encoder-decoder Transformer, potentially correcting errors and adding complexity to the representation. The first block is connected through a skip connection that extends beyond the subsequent Transformer block, directly interfacing with the convolutional (Conv) decoder. The second block also utilizes a skip connection, contributing its processed features directly to the Conv decoder. The third block follows a direct sequential path, transmitting its output to the Conv decoder. The skip connections allow information to be passed directly from one layer to another, allowing the network to effectively learn complex relationships in the data while mitigating the risk of vanishing gradients \cite{oord2016wavenet, he2016deep}.} The training process is outlined in Algorithm \ref{alg:training}.
\begin{algorithm}
	\renewcommand{\algorithmicrequire}{\textbf{Input:}}
	\renewcommand{\algorithmicensure}{\textbf{Output:}}
	\caption{Training}
	\begin{algorithmic}
        \REQUIRE $\mathbf{x}_0^{ta}\sim q(\mathbf{x}_0^{ta}|\mathbf{x}_0^{co})$, $\mathbf{x}_0^{co}$
        \ENSURE $\epsilon _\theta$ network trained
        \REPEAT
        \STATE $t\sim Uniform({1,...,T})$
        \STATE $\epsilon \sim N(\mathbf{0},\mathbf{I})$
        \STATE Take gradient descent step on $\nabla _\theta \Vert \epsilon -\epsilon _\theta (\mathbf{x}_t^{ta},t|\mathbf{x}_0^{co})\Vert _2 ^2$
		\UNTIL Converged
	\end{algorithmic}  
	\label{alg:training}
\end{algorithm}

\subsection{\textcolor{black}{Model Inference}}
The values of the predicted triplets in the model are initially Gaussian noise corresponding to the first diffusion step of the reverse trajectory (step $T$ in Figure \ref{fig:diffusion_process}). These noisy values are then denoised using the output from $\epsilon _\theta$ according to Equation \ref{eq:reverse}, which is written as $\mathbf{x}_{t-1}^{p}=\frac{1}{\sqrt{\hat{\alpha}}_t}({x}_t^{p}-\frac{\beta _t}{{\sqrt{1-\alpha}_t}}\epsilon _\theta(x_t^{p},t|x_0^{co}))+\sigma _t\mathbf{z}$, where $\mathbf{z} \sim N(\mathbf{0},\mathbf{I})$. The features and times of the triplets to predict provide important contextual information. The denoised values are then fed back into the model to generate a prediction of $\epsilon$ for the next diffusion step, and this process is repeated until the final step of the reverse trajectory to output $\mathbf{x}_0^p$. The detailed process for this inference is described in Algorithm \ref{alg:inference}.
\begin{algorithm}
	\renewcommand{\algorithmicrequire}{\textbf{Input:}}
	\renewcommand{\algorithmicensure}{\textbf{Output:}}
	\caption{Inference}
	\begin{algorithmic}
        \REQUIRE $\mathbf{x}_T^{p}\sim N(\mathbf{0},\mathbf{I})$, $\mathbf{x}_0^{co}$
        \ENSURE $\mathbf{x}_0^{p}$ (prediction of $\mathbf{x}_0^{ta}$)
        \FOR{$t=T,...,2$}
            \STATE $\mathbf{z} \sim N(\mathbf{0},\mathbf{I})$
            \STATE $\sigma _t^2=\frac{1-\alpha _{t-1}}{1-\alpha _t}\beta _t$
            \STATE $\mathbf{x}_{t-1}^{p}=\frac{1}{\sqrt{\hat{\alpha}}_t}({x}_t^{p}-\frac{\beta _t}{{\sqrt{1-\alpha}_t}}\epsilon _\theta(x_t^{p},t|x_0^{co}))+\sigma _t\mathbf{z}$
        \ENDFOR
            \STATE $\mathbf{x}_0^{p}=\frac{1}{\sqrt{\hat{\alpha}}_1}({x}_1^{p}-\frac{\beta _1}{{\sqrt{1-\alpha}_1}}\epsilon _\theta(x_1^{p},1|x_0^{co}))$
	\end{algorithmic}
	\label{alg:inference}
\end{algorithm}
\section{Experiments}
\subsection{Data and Preprocessing}
\begin{figure}
\centering
\includegraphics[width=.9\textwidth]{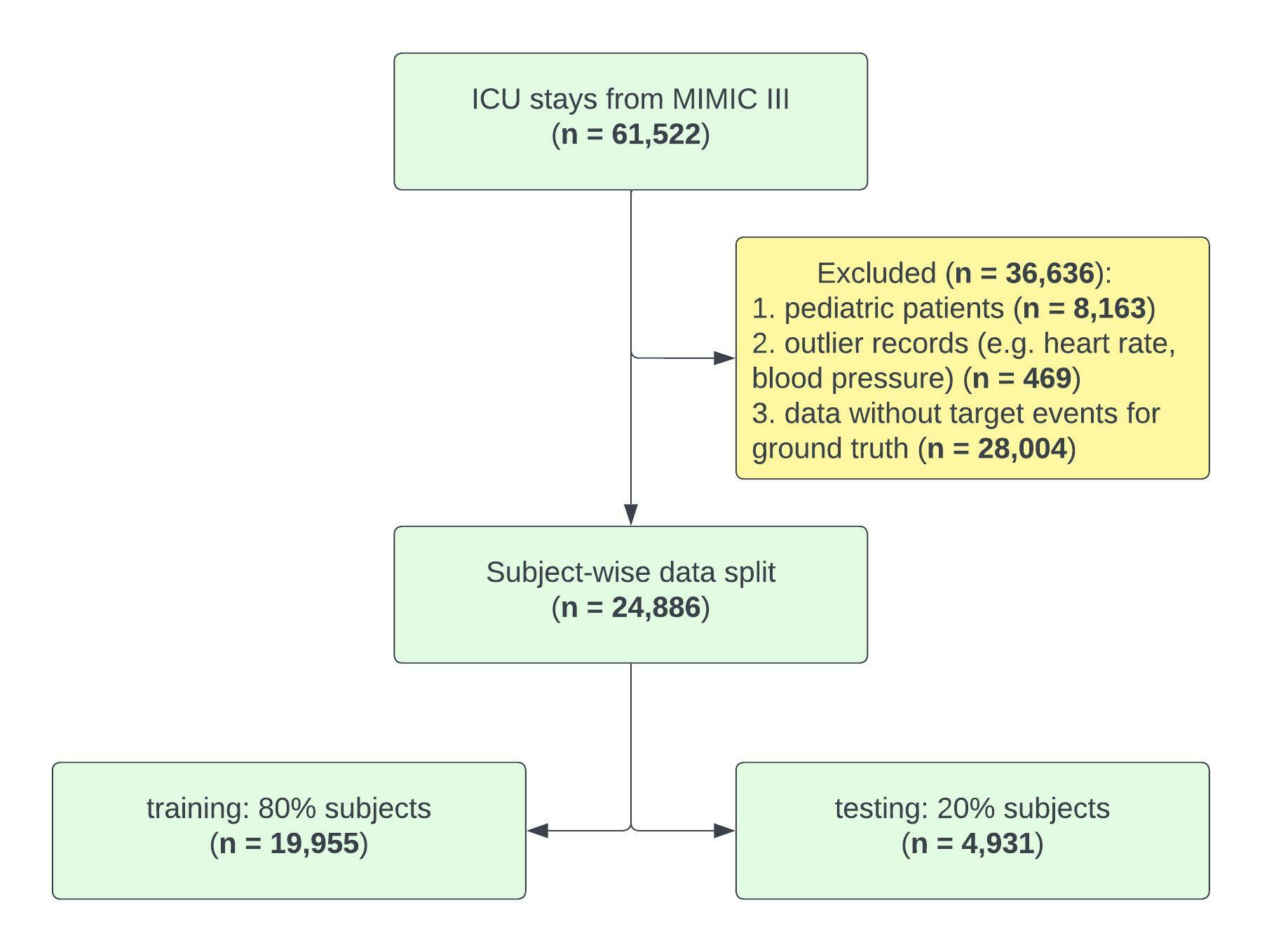}
\caption{Illustration of the data preprocessing steps involved in this study. $n$ denotes the number of ICU stays.}
\label{fig:data_preprocessing}
\end{figure}

In this study, we evaluate the model using the MIMIC-III dataset \cite{johnson2016mimic}. This dataset holds health information for over $46$ thousand patients admitted to the Beth Israel Deaconess Medical Center (Boston, MA) between $2001$ and $2012$. \textcolor{black}{The data preprocessing approach is depicted in Figure \ref{fig:data_preprocessing}. The $3$ steps in the yellow box are detailed as follows.} First, we exclude records related to pediatric patients \textcolor{black}{by filtering out individuals whose ages are greater than or equal to $18$ years old.} Second, we remove records with abnormal feature values. The features represent the medical events that happen to the patients, such as vital signs, medication usage, and biochemical test results. \textcolor{black}{We eliminate outliers by retaining values within reasonable ranges for $117$ numerical features, while non-null values are retained for other features categorized as yes-or-no items.} Ultimately, we solely preserve the initial $40$ consecutive minutes from each ICU stay, denoting this piece of data as an individual ICU sample. Of these $40$ minutes, the former $30$ minutes are used for the conditional data, and the latter $10$ minutes for the target data. Both the conditional and target data are screened to ensure non-emptiness. \textcolor{black}{We exclude ICU samples where the latter $10$-minute segment lacks any target data. We standardize the feature values to ensure that it is suitable for input into the TDSTF model. We partially adopt the excluding method from \cite{tipirneni2022self} to steps $1$ and $2$.} To prevent subject repetition, we divide the dataset into training and testing sets, randomly selecting $80\%$ and $20\%$ of the data by subject. 

After implementing the exclusion criteria in the preprocessing step, a total of $11,401$ subjects were included in the study. From these eligible subjects, we extracted $24,886$ ICU stays to test our method. Table \ref{tab:patient} presents statistics for the eligible subjects and target features in both the training and testing datasets before standardization. Subjects in the training and testing datasets combined were generally older (age of $65.42\pm 16.27$) and male ($57.2\%$). The training data is further split into an $80-20\%$ ratio, with the latter portion used for validation. Note that one subject corresponds to one or more ICU stays. The goal of this study is to predict three target features: HR,  SBP, and DBP \textcolor{black}{\cite{lockwood2004vital}}.
\begin{table}
\caption{Subject number by sex; mean and standard deviation of subject age and target features.}
\label{tab:patient}
\begin{center}
\begin{tabular}{ccc}
\toprule
&Training&Testing\\
&n=19,955&n=4,931\\
\midrule
M/F&5244/3919&1281/957\\
Age&65.43$\pm$16.35&65.36$\pm$15.94\\
HR&90.82$\pm$21.56&91.24$\pm$22.27\\
SBP&117.74$\pm$27.95&117.76$\pm$30.45\\
DBP&60.34$\pm$16.86&59.90$\pm$17.05\\
\bottomrule
\end{tabular}
\end{center}
HR is heart rate in beats per minute (bpm); SBP is systolic blood pressure in millimeters of mercury (mmHg); DBP is diastolic blood pressure in mmHg; M is male; F is female; $n$ is the number of ICU stays.
\end{table}

\subsection{Metrics}
To thoroughly assess the performance of the proposed method, multiple metrics are utilized to measure the differences between the forecast and the ground truth time series.

One of the simplest and most commonly used metrics for this purpose is the Mean Squared Error (MSE). It is defined as the squared $L2$ distance between each element in the input $\mathbf{x}$ and target $\mathbf{y}$, calculated as
\begin{align}
MSE(\mathbf{x},\mathbf{y}) = \sum_i (x_i-y_i)^2
\end{align}

In our case, the median values of the generated samples act as the deterministic predictions for the calculation of MSE with the ground truth. A lower MSE value indicates better performance.

\begin{figure}
\centering
\includegraphics[width=.9\textwidth]{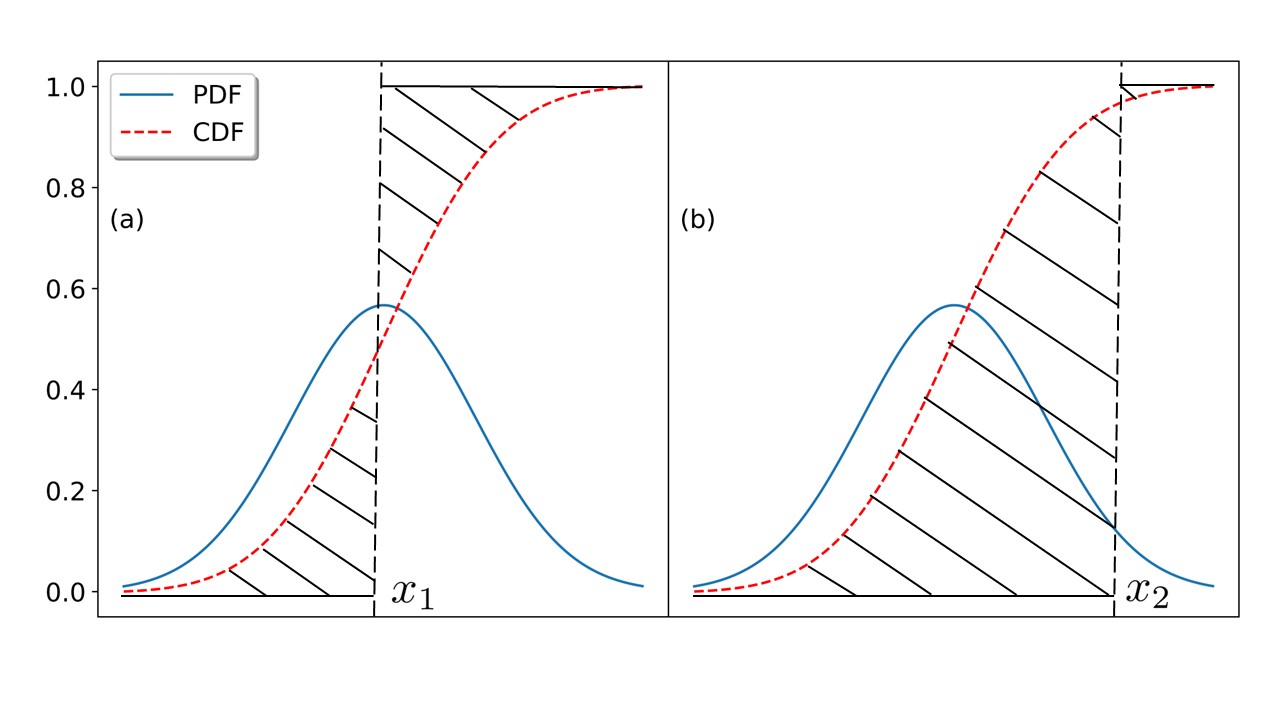}
\caption{A schematic illustration of the Continuous Ranked Probability Score (CRPS). The blue curves represent the modeled probability distribution functions, and the red dashed curves represent their cumulative distribution functions. The subplots (a) and (b) depict the same distribution with different ground truth $x_1$ and $x_2$. The black hatched area representing the integration as the CRPS value in (a) is smaller than that in (b), meaning the modeled distribution evaluates $x_1$ better.}
\label{fig:crps}
\end{figure}

Another widely used metric for evaluating the accuracy of probabilistic forecasts is the Continuous Ranked Probability Score (CRPS). It is defined as:
\begin{align}
CRPS(F,x)=\int _{-\infty} ^\infty (F(y)-I_{y\geq x})^2dy
\end{align}
\noindent where $F$ is the modeled Cumulative Distribution Function (CDF), $x$ is the observation, and $I$ is the Indicator function. Figure \ref{fig:crps} illustrates how CRPS is calculated. It provides a comprehensive measure of forecast accuracy by evaluating the mean difference between the predicted CDF and the ground truth, taking into account both under-prediction and over-prediction. The smaller the CRPS value, the more the modeled distribution concentrates around the ground truth, indicating better performance.

To calculate the \textcolor{black}{Standardized} Average CRPS (SACRPS), when predicting a \textcolor{black}{standardized} target value $x^{ta}$, we adopt the same discrete quantile loss as in \cite{tashiro2021csdi}:
\begin{align}
&SACRPS=\sum_{x^{ta}}\sum_{i=1}^{19}2\Lambda _{0.05i}(x_{0.05i}^p,x^{ta})/19\sum |x^{ta}|\\
&\Lambda _\alpha(q,x)=(\alpha -I_{q\geq x})(x-q)
\end{align}
\noindent where $x_{0.05i}^p$ is the value of $0.05i$ quantile level from the generated samples. The scaling factor $\sum |x^{ta}|$ represents the variance of the target distribution, which becomes more difficult to predict as the variance increases. \textcolor{black}{SACRPS} calculated as the CRPS divided by this scaling factor provides a more accurate evaluation of the model.

\subsection{Experiment Setup}
 We use the ADAM optimizer \cite{kingma2014adam} with a learning rate $10^{-3}$. The mini-batch size is set to $32$ samples. We implemented all the experiments using the Python programming language and the Pytorch framework \cite{paszke2019pytorch}. The experiments were conducted on a workstation that was equipped with an Intel i$7$-$12700$K processor and an Nvidia RTX $3090$ graphics card.

\begin{figure}
\centering
\includegraphics[width=.9\textwidth]{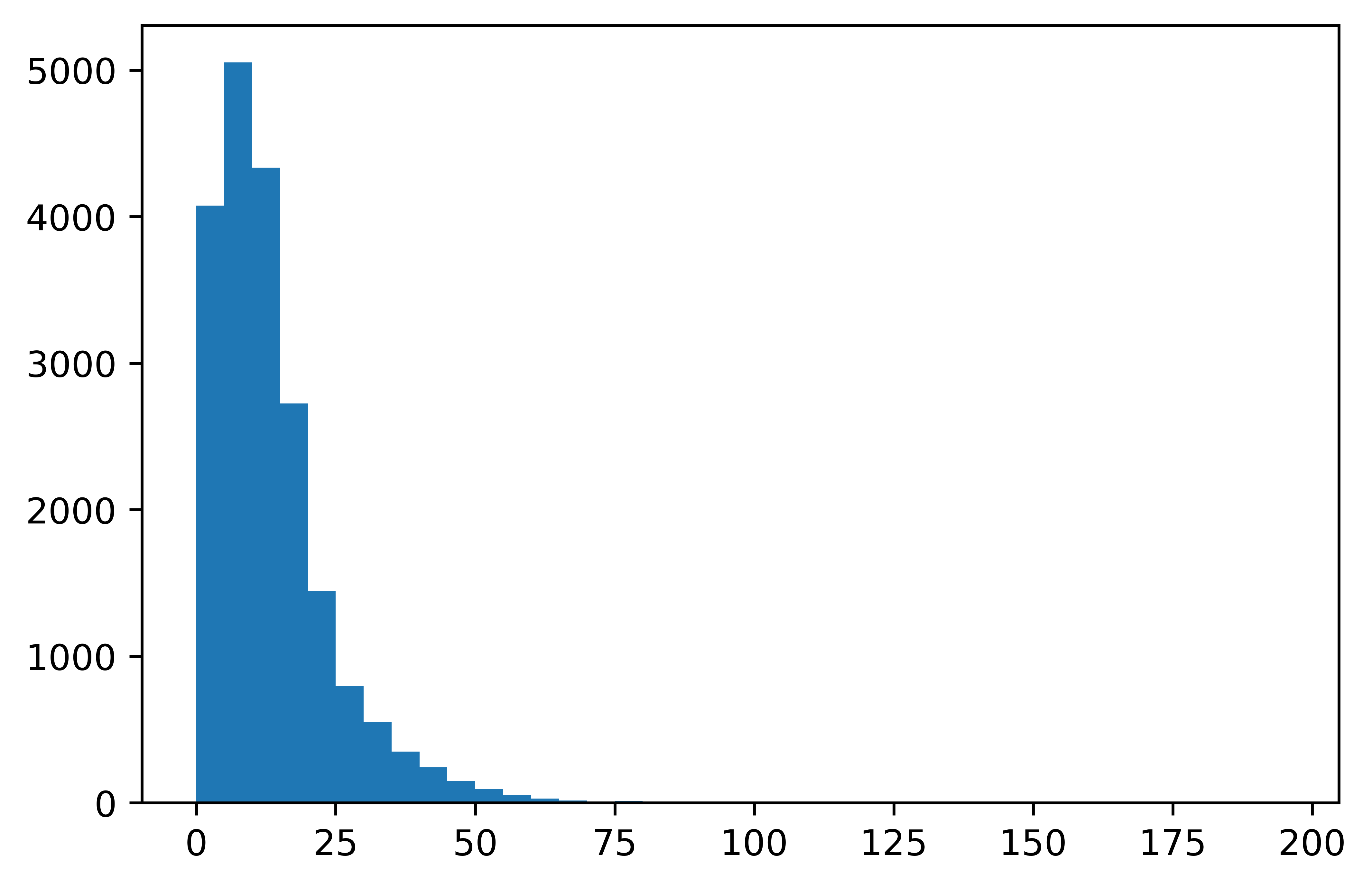}
\caption{Histogram showing the distribution of valid conditional data points derived from ICU samples within the training dataset, exhibiting a mean count of $12.9$. On average, $99.7\%$ of the conditional data within the matrix representation exhibit missing values. More than $99.5\%$ of the ICU samples from the training dataset contain less than or equal to $60$ valid conditional data points.}
\label{fig:valid_points}
\end{figure}

\textcolor{black}{\subsection{Comparison with Baseline}}
Five distinct models were employed as the baseline models for predicting vital signs with the MIMIC-III dataset. \textcolor{black}{MQ-RNN encompasses a singular layer of bidirectional GRU, featuring a hidden size of $50$. The encoder CNN employs kernel sizes of $[7,3,3]$ alongside channel counts of $[30,30,30]$. Each layer within the Forking MLP Decoder is dimensioned at $30$. The lookback period is tailored to cover the entire input time length. Categorical feature embeddings assume a dimensionality of $50$. DeepAR comprises $2$ layers of LSTM, each with $40$ cells. The Dimension of the embeddings for categorical features is set as $50$. The context length is adjusted to encompass the prediction horizon. For evaluation and sampling predictions, the student-t distribution is employed. DeepFactor applies both global and local models instantiated as LSTM. The count of global factors stands at $10$, with each global model housing a single hidden layer of $50$ units. The local model comprises a hidden layer containing $5$ units. The observation model is specified as student-t. EnCQR contains an ensemble of $5$ member learners, each configured as a $5$-layer LSTM with a hidden size of $128$ and an $L2$ regularization factor of $10^{-4}$. Conformalized prediction intervals are utilized as the output. CSDI involves $50$ diffusion steps, with a noise schedule following the quadratic schedule from $\beta_1=10^{-4}$ to $\beta_{50}=0.5$ with a quadratic diffusion schedule \cite{tashiro2021csdi}. Feature, time, and diffusion embedding dimensions are established at $16$, $128$, and $128$, respectively.} To address the missing data, we adopted a mean imputation of $0$ for MQ-RNN, DeepAR, DeepFactor, and EnCQR, whereas CSDI utilized a mask matrix to denote the positions of missing data. In our model, the input size of the conditional triplet array is set to $60$ according to Figure \ref{fig:valid_points}.

\begin{figure}
\centering
\includegraphics[width=.9\textwidth]{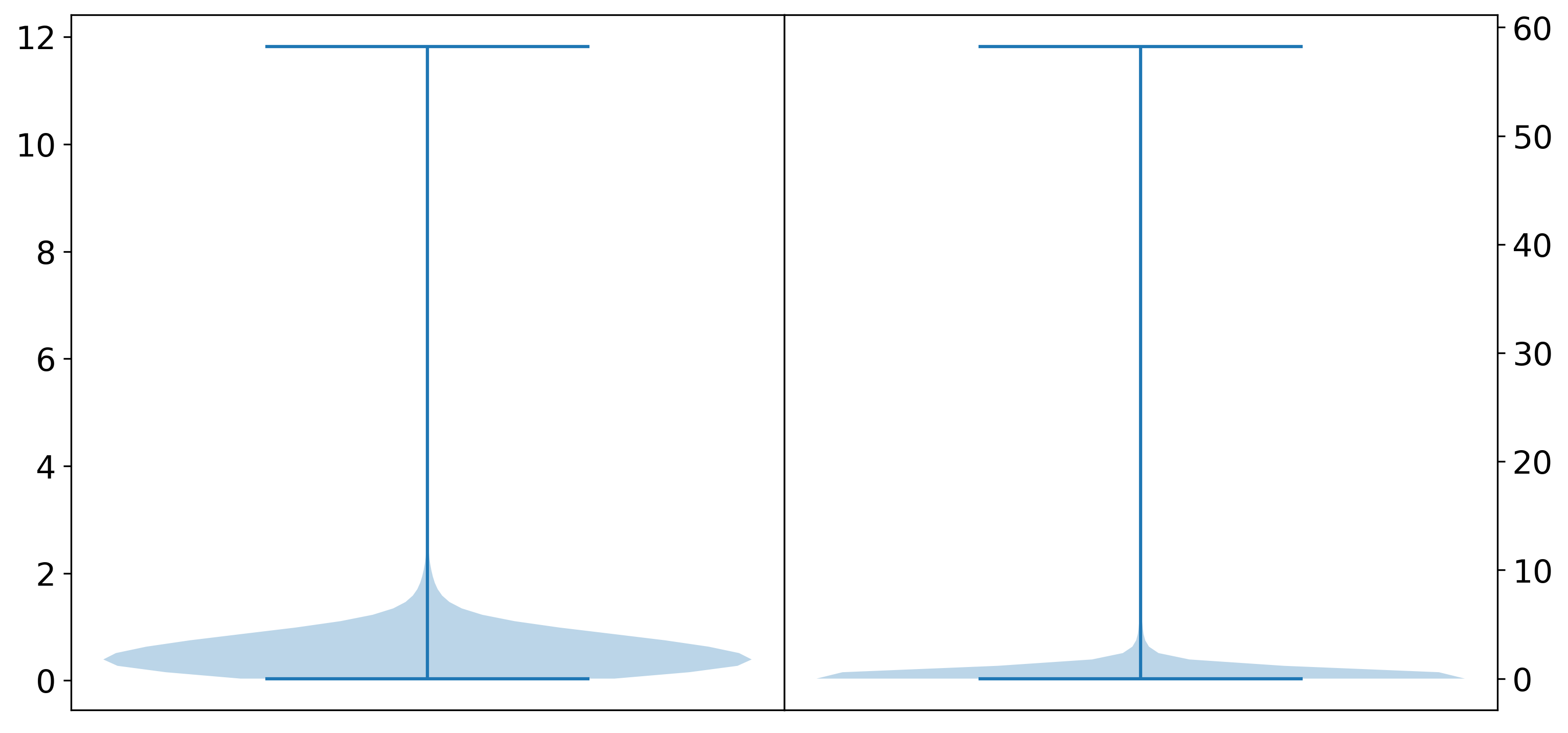}
\caption{Violin histograms of the \textcolor{black}{SACRPS} in the left and the MSE in the right per ICU sample from the first trial of TDSTF. The results for each metric concentrate around the median values of $0.4651$ and $0.1677$.}
\label{fig:per_sample}
\end{figure}

The experimental results are presented in Table \ref{tab:baseline}, with $3$ trials conducted. Both the \textcolor{black}{SACRPS} and MSE results indicate that the CSDI and TDSTF models performed much better than the other models. On average, TDSTF obtained $0.4438$ and $0.4168$, while CSDI obtained $0.5439$ and $0.6358$ for \textcolor{black}{SACRPS} and MSE, respectively. Thus, TDSTF improved \textcolor{black}{SACRPS} and MSE by $18.9\%$ and $34.3\%$ over CSDI. Figure \ref{fig:per_sample} shows the violin histograms of \textcolor{black}{SACRPS} and MSE per ICU sample for the first trial of TDSTF. The results concentrate on the median values $0.4651$ and $0.1677$ of \textcolor{black}{SACRPS} and MSE, which shows robustness of the model.
\begin{table}
\caption{\textcolor{black}{SACRPS} and MSE of the proposed model in comparison with the baselines. The best results are shown bold.}
\label{tab:baseline}
\begin{center}
\begin{tabular}{ccccccc}
\toprule
&MQ-RNN&DeepAR&DeepFactor&\textcolor{black}{EnCQR}&CSDI&TDSTF (Ours)\\
\midrule
\textcolor{black}{SACRPS}&1.0276&0.9930&0.8221&\textcolor{black}{0.8255}&0.5515&\textbf{0.4379}\\
&1.0367&0.9615&0.8207&\textcolor{black}{0.8256}&0.5455&\textbf{0.4526}\\
&1.0385&0.9653&0.8238&\textcolor{black}{0.8254}&0.5439&\textbf{0.4408}\\
\midrule
MSE&1.1014&1.0295&1.0325&\textcolor{black}{1.2711}&0.6430&\textbf{0.4061}\\
&1.1237&1.0304&1.0332&\textcolor{black}{1.2710}&0.6358&\textbf{0.4434}\\
&1.1186&1.0289&1.0324&\textcolor{black}{1.2715}&0.6236&\textbf{0.4008}\\
\bottomrule
\end{tabular}
\end{center}
\end{table}

\begin{figure}
    \centering
    \subfigure[An increase in HR.]{
        \includegraphics[width=.27\textwidth]{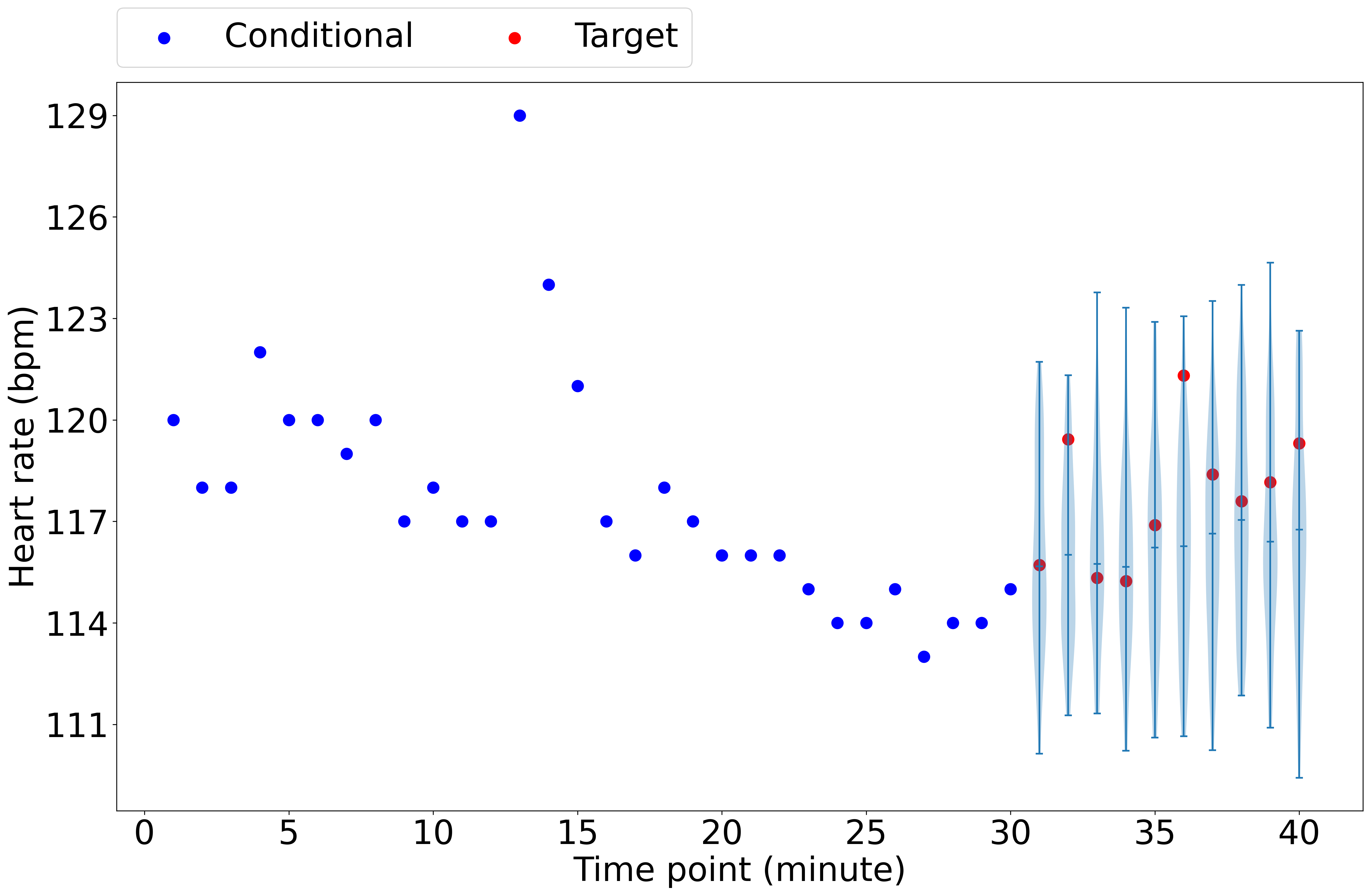}
    }
    \subfigure[A sudden increase in HR.]{
        \includegraphics[width=.27\textwidth]{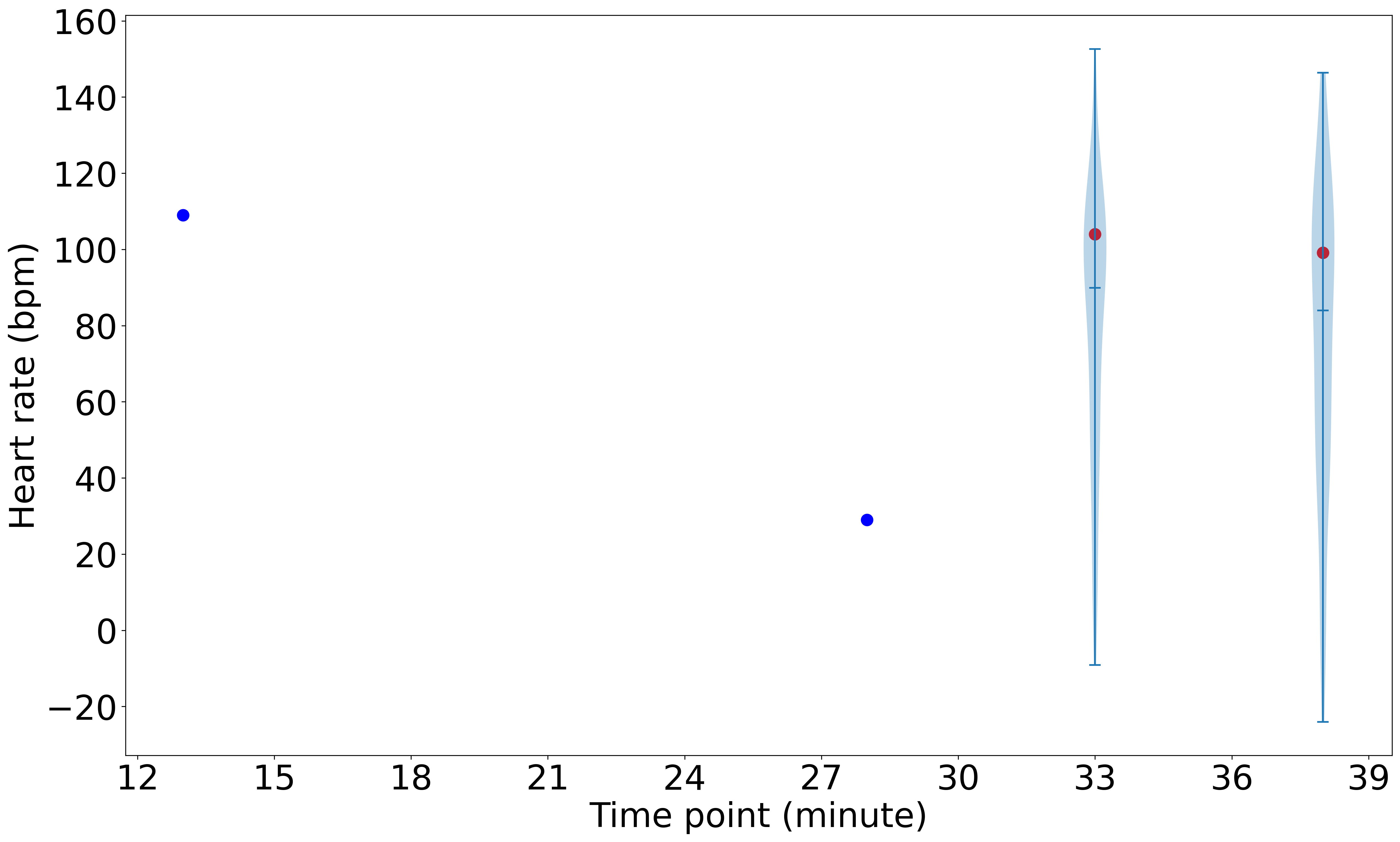}
    }
    \subfigure[HR prediction without HR conditional data.]{
        \includegraphics[width=.27\textwidth]{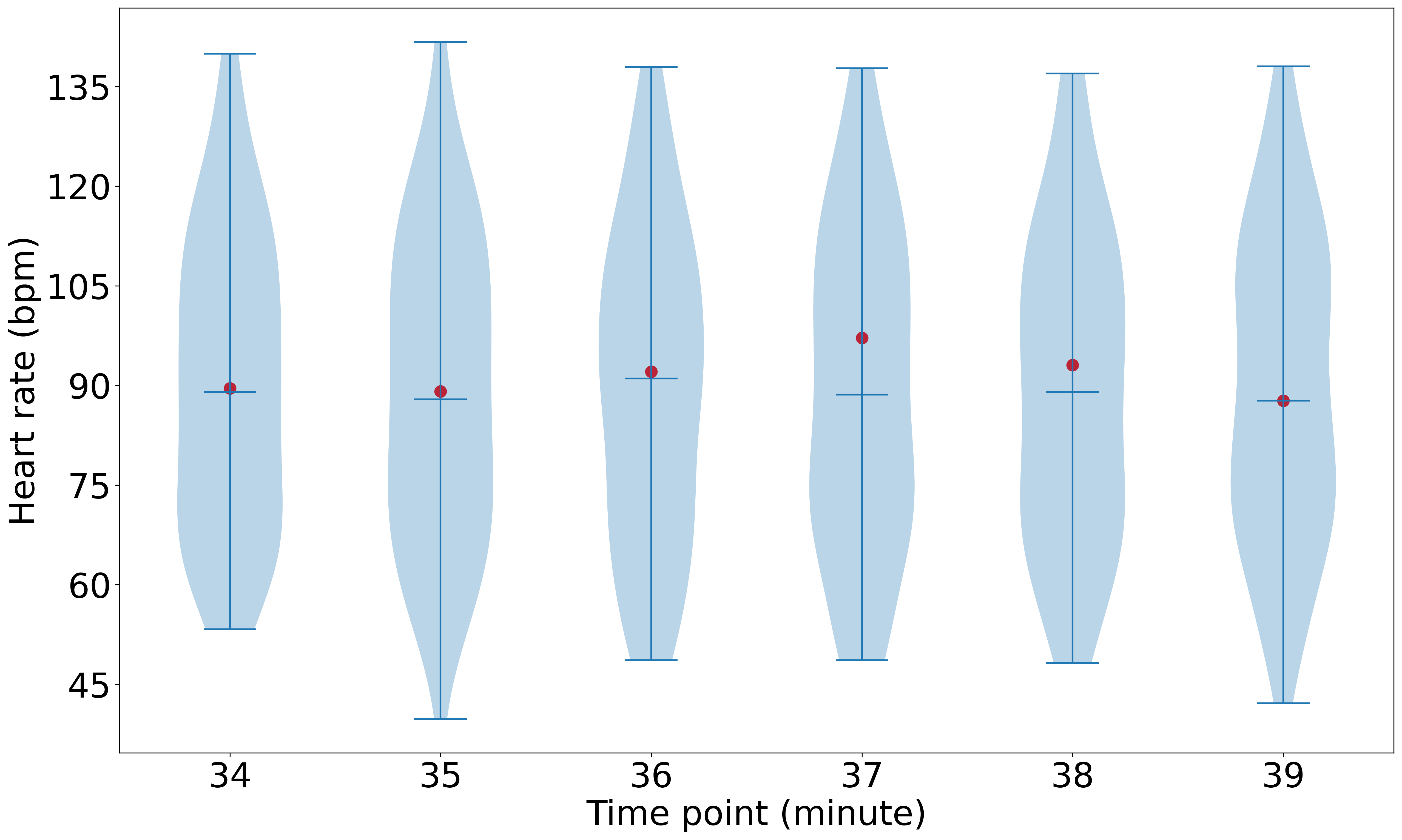}
    }
    \subfigure[A decrease in SBP.]{
        \includegraphics[width=.27\textwidth]{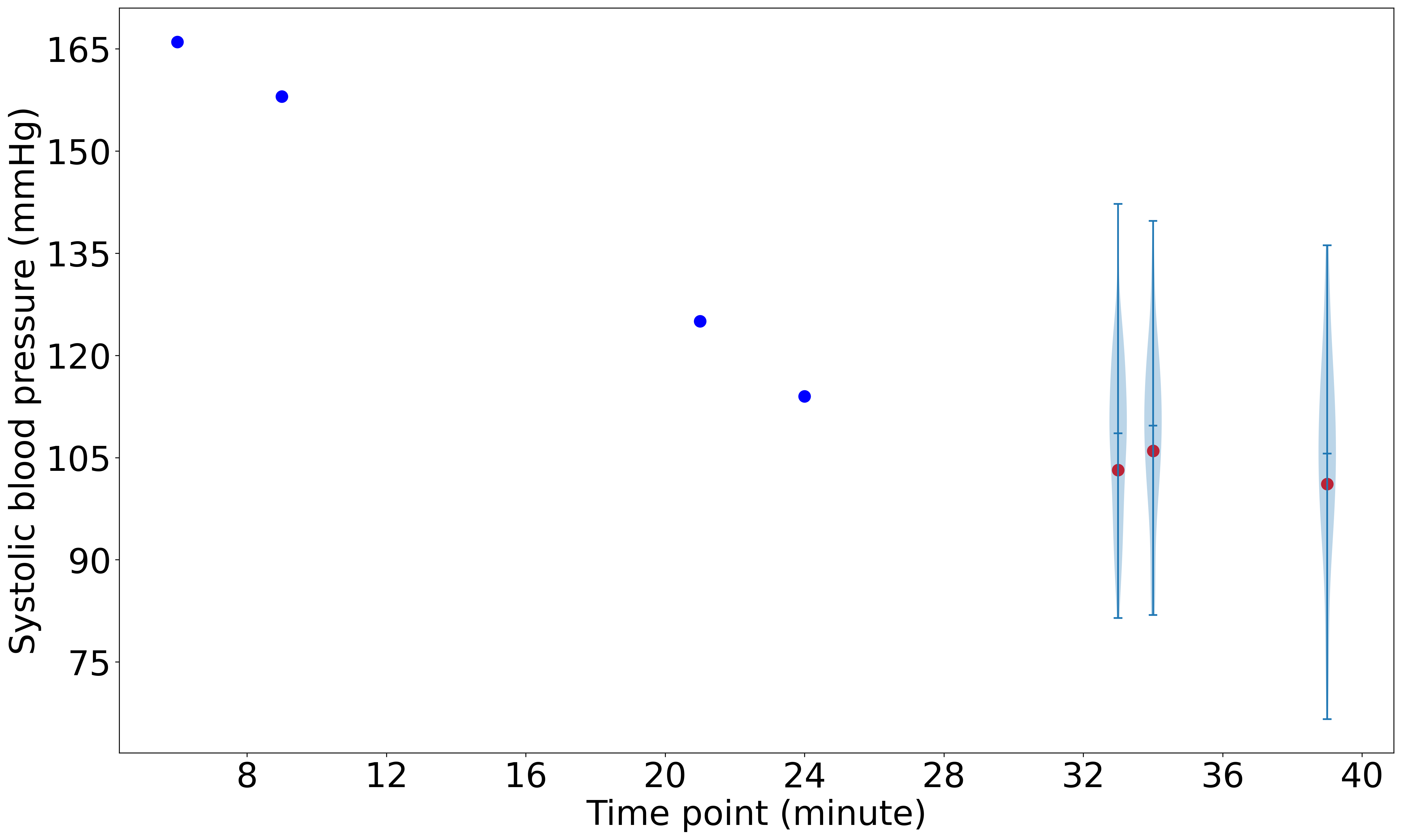}
    }
    \subfigure[A sudden decrease in SBP.]{
        \includegraphics[width=.27\textwidth]{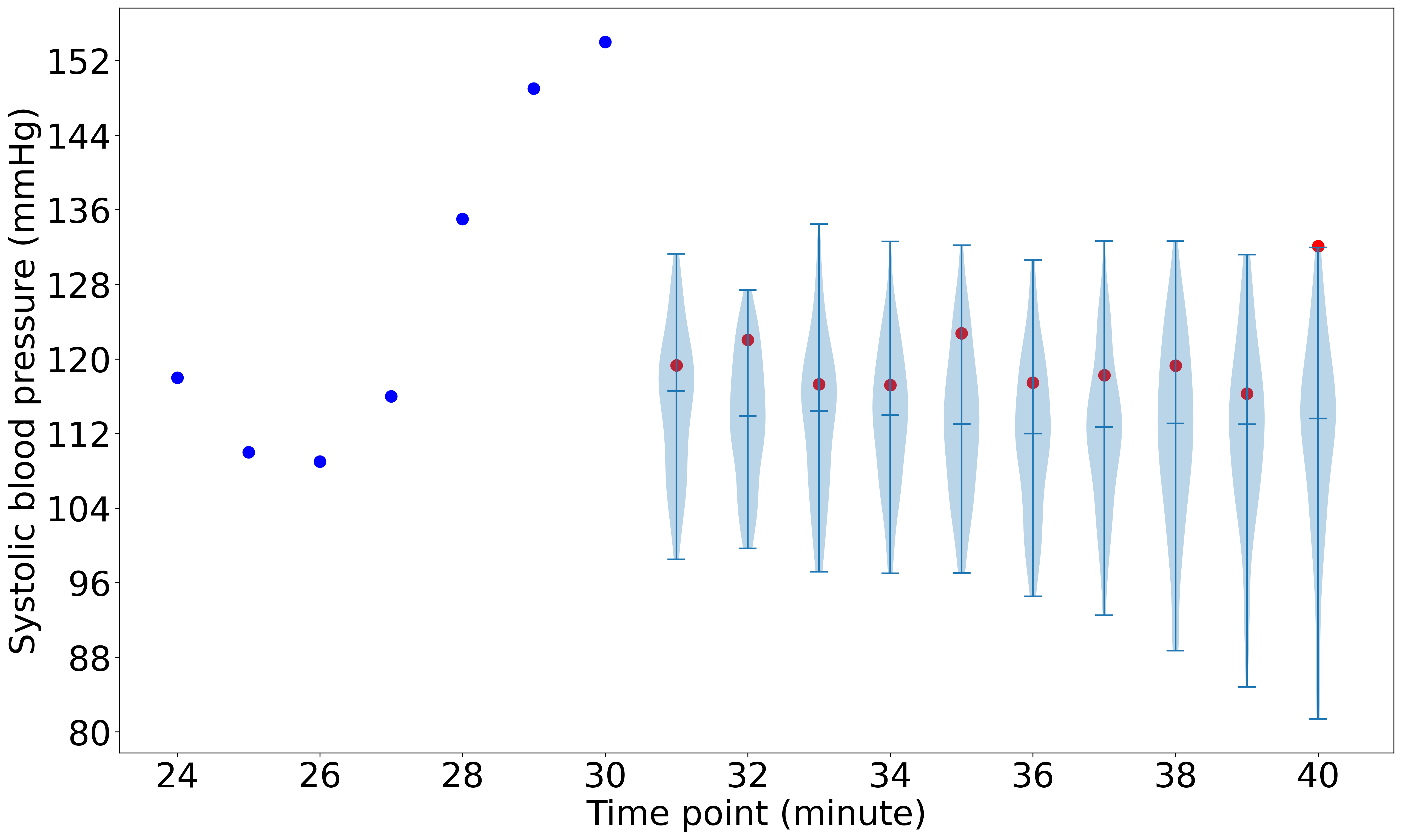}
    }
    \subfigure[SBP prediction without SBP conditional data.]{
        \includegraphics[width=.27\textwidth]{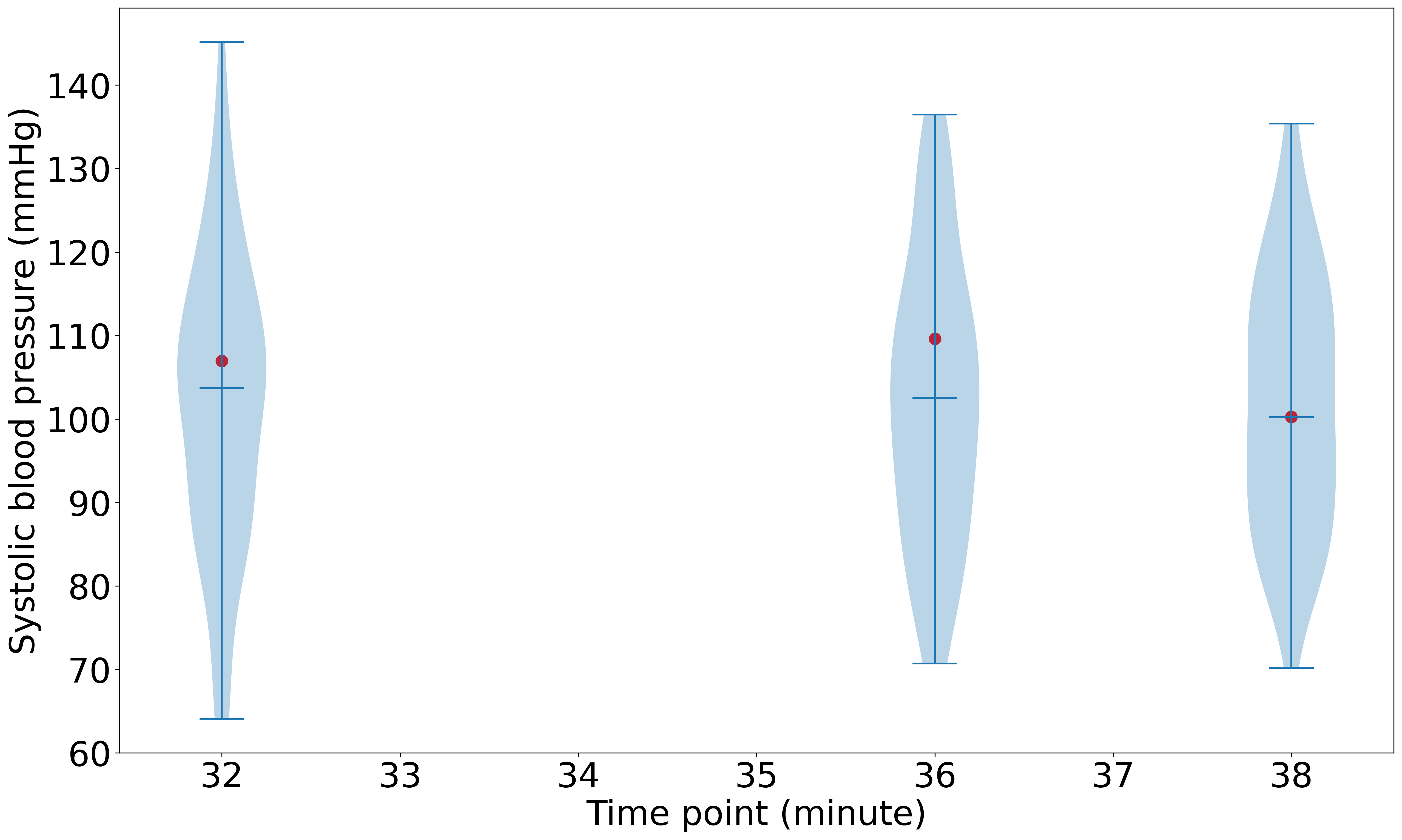}
    }
    \subfigure[A decrease in DBP.]{
        \includegraphics[width=.27\textwidth]{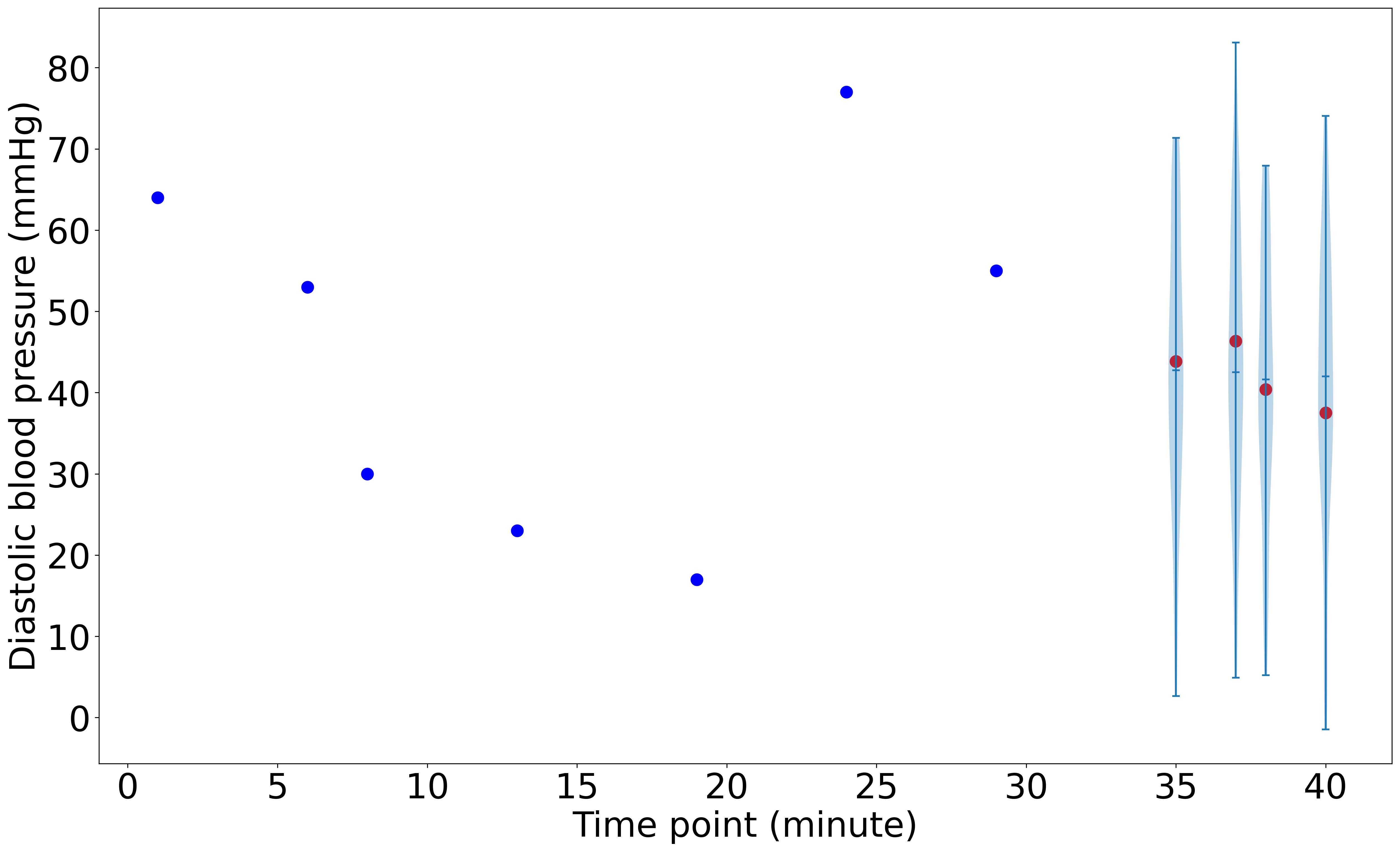}
    }
    \subfigure[A sudden decrease in DBP.]{
        \includegraphics[width=.27\textwidth]{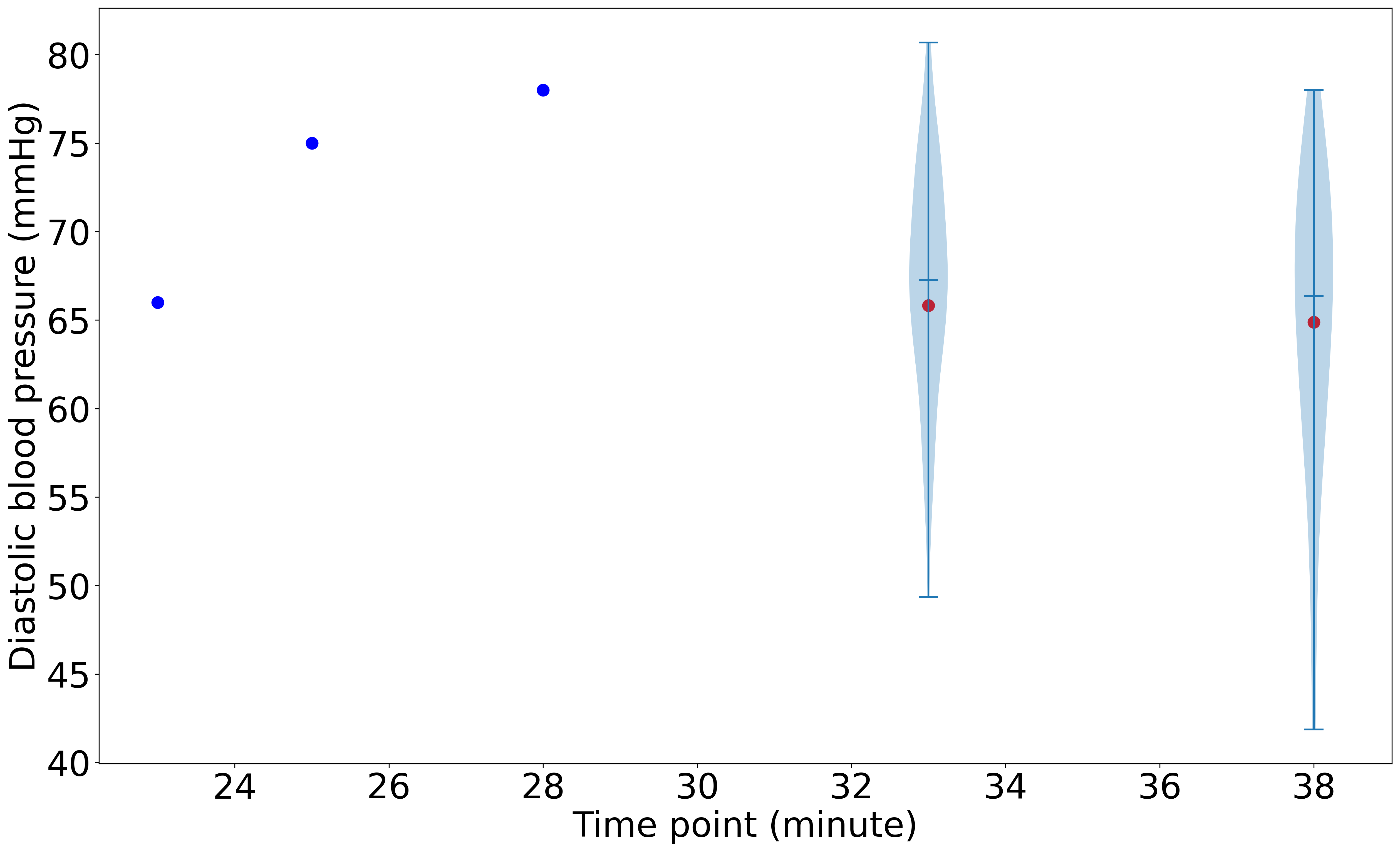}
    }
    \subfigure[DBP prediction without DBP conditional data.]{
        \includegraphics[width=.27\textwidth]{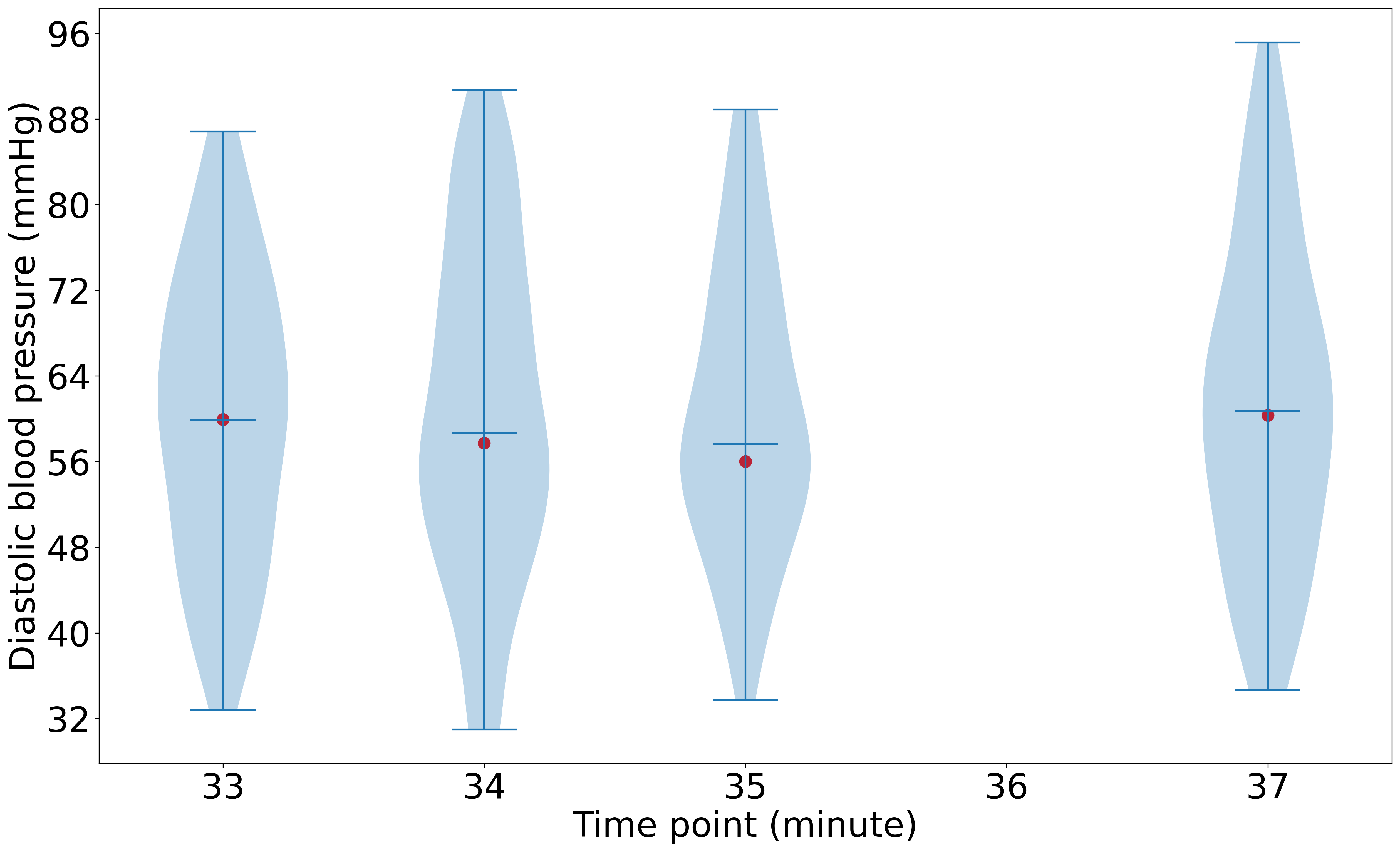}
    }
    \caption{Examples from the TDSTF test results from the first trial of TDSTF. HR predictions in subplots (a)-(c) are in beats per minute (bpm), while SBP and DBP predictions in subplots (d)-(i) are in millimeters of mercury (mmHg). The horizontal axis represents the relative time in minutes for each ICU stay. The red dots indicate target feature values, and the blue dots indicate conditional values of the same target feature. Forecasts based on all $129$ features show high accuracy, even without the conditional data as is shown in (c), (f), and (i). The $95\%$ confidence intervals are shown as the areas between top and bottom bars of the violin histograms, with wider areas corresponding to more confidence and the middle bars representing the median values of the generated data points.}
    \label{fig:examples}
\end{figure}

Figure \ref{fig:examples} shows some forecasting examples from the first trial of TDSTF. All the deterministic predictions fall into the $95\%$ confidence interval. Subplot (a) predicts an increasing trend of HR over $100$ beats per minute (bpm). A sudden increase in HR to around $100$ bpm is captured in subplot (b). Both of them imply potential tachycardia. Subplot (d) shows a continuously decreasing SBP, and possible hypotension may be expected. Subplot (e) forecasts recovery from hypertension into a period of a normotensive SBP of around $120$ millimeters of mercury (mmHg). A decreasing trend of DBP below $60$ mmHg is seen in subplot (g), suggesting the patient is at risk for dangerously low blood pressure. Subplot (h) captures a sudden decrease in DBP, which should alert clinicians of the potential for further hypotension. The model also performed well on test cases without target conditional data, as is shown in subplots (c), (f), and (i).

The complexity of TDSTF was compared with CSDI using PyTorch-OpCounter \footnote{\url{https://github.com/Lyken17/pytorch-OpCounter}}. The comparison considered the number of parameters and the Multiply-Accumulate (MAC) of the models. The number of parameters determines the model complexity, while the MAC number refers to the computational costs. The results show that CSDI had $0.28$ million parameters, while TDSTF had $0.56$ million parameters. The training time for CSDI was $12$ hours, while for TDSTF, it was only $0.5$ hours. The inference per sample consumed more than $55$ billion MACs in CSDI and $0.79$ million in TDSTF. Consequently, $14.913$ and $0.865$ seconds were required for inference per sample in CSDI and TDSTF, respectively, and TDSTF was more than $17$ times faster. The fact that TDSTF is more complex than CSDI but consumes much less computation verifies our assumption on the CSDI overhead due to its large amount of missingness input.

\begin{figure}
\centering
\includegraphics[width=.9\textwidth]{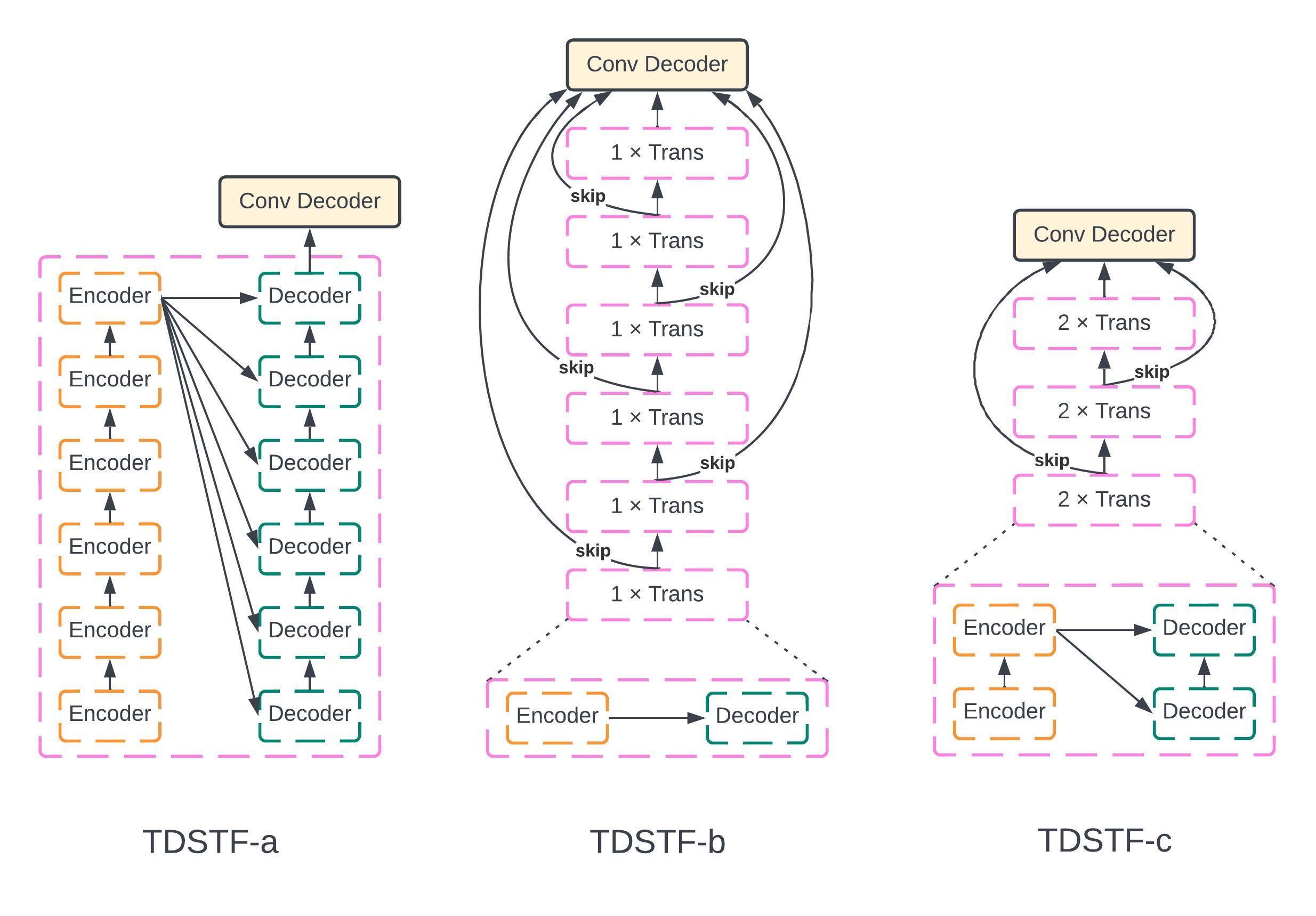}
\caption{The backstage structures for the ablation study.}
\label{fig:ablation}
\end{figure}

\textcolor{black}{\subsection{Ablation Study}}
\textcolor{black}{To assess the impact of Transformer block structures on the backbone's architecture and the diffusion model’s performance, we conducted an ablation study. This study focused on $3$ configurations, as delineated in the \ref{fig:ablation}. With a series of $3$ trials, the outcomes are shown in Table \ref{tab:ablation}. TDSTF-a and TDSTF-c require the same training duration of 0.5 hours. TDSTF-c demonstrates better performance and robustness compared to TDSTF-a. Although TDSTF-b achieves lower MSE than TDSTF-c, it shows higher SACRPS and demands a longer training time of 0.7 hours. However, TDSTF-c manages to maintain comparable overall performance with a shorter training duration. As a result, TDSTF-c achieves an optimal balance between performance and model efficiency.}

\begin{table}
\caption{SACRPS and MSE of the $3$ ablation configurations, each with $\beta _1=10^{-4}$ and $\beta _T=0.5$. The best results are shown in bold. TDSTF-c is highlighted as the selected configuration.}
\label{tab:ablation}
\begin{center}
\begin{tabular}{cccc}
\toprule
Configuration&TDSTF-a&TDSTF-b&\textbf{TDSTF-c}\\
\midrule
SACRPS&0.4843&\textbf{0.4328}&0.4379\\
&0.6072&0.4720&\textbf{0.4526}\\
&0.5162&0.4442&\textbf{0.4408}\\
\midrule
MSE&0.7861&\textbf{0.3188}&0.4061\\
&1.2609&\textbf{0.4247}&0.4434\\
&0.6341&\textbf{0.3506}&0.4008\\
\bottomrule
\end{tabular}
\end{center}
\end{table}

\textcolor{black}{Noise amount plays a critical role in diffusion models, thus prompting our investigation of the $\beta$ schedule through additional $2$ sets of experiments, each consisting of $3$ trials, built upon TDSTF-c. In both scenarios, $\beta _1$ remains fixed at $10^{-4}$, and $T$ is still set to $50$. $\beta _T$ assumes values of $0.05$ and $0.1$ respectively. The results in Table \ref{tab:beta} indicate that the schedule spanning from $\beta _1=10^{-4}$ to $\beta _T=0.5$ yields the most optimal performance.}
\begin{table}
\caption{SACRPS and MSE of Configuration c with $3$ $\beta_T$ values. The best results are shown in bold.}
\label{tab:beta}
\begin{center}
\begin{tabular}{cccc}
\toprule
$\beta_T$&0.05&0.1&0.5 (Ours)\\
\midrule
SACRPS&0.5898&0.4893&\textbf{0.4379}\\
&0.5884&0.4721&\textbf{0.4526}\\
&0.6002&0.5023&\textbf{0.4408}\\
\midrule
MSE&0.5247&0.4613&\textbf{0.4061}\\
&0.5629&0.4443&\textbf{0.4434}\\
&0.5777&0.4847&\textbf{0.4008}\\
\bottomrule
\end{tabular}
\end{center}
\end{table}
\section{Discussion}
This study presents a deep learning model, TDSTF, that can accurately predict sparse time series data in the ICU setting. The model was trained on MIMIC-III data and tested on several performance metrics. The results show that TDSTF outperforms several baseline models and can detect important changes in the vital signs of ICU patients. The model consists of a residual network based on Transformers, which allows for the ability to capture complex temporal dependencies and efficient computation and storage.

The TDSTF model is developed to accurately forecast sparse time series data without making assumptions about the underlying distribution. The model was trained and tested using the MIMIC-III ICU data and evaluated using \textcolor{black}{SACRPS} and MSE. It outperforms the baseline models, including MQ-RNN, DeepAR, DeepFactor, EnCQR, and CSDI. The mask used in CSDI reduces disturbance but cannot guarantee enough exclusion of invalid information from the sparse time series input. On average, $99.7\%$ of conditional data points were missing, which could negatively impact performance and waste computational resources. Setting the input size of conditional triplets to $60$ for TDSTF improved the signal-to-noise ratio and outperformed CSDI because of its improved representational power. MQ-RNN, DeepAR, DeepFactor, and EnCQR use RNN to extract hidden states and cannot fully utilize parallel computation, hindering model complexity \cite{de2015survey}. The Transformer-based $\epsilon_\theta$ network solves these issues. While DeepAR and DeepFactor \textcolor{black}{assume likelihood that may be} too simple to represent high-dimensional distributions. TDSTF, on the other hand, directly learns the underlying distributions.

TDSTF is effective because it accurately captures temporal patterns in sparse time series data, as shown in Figure \ref{fig:examples}. In addition to high accuracy in the slow change cases, the model successfully recognizes sudden changes, which are important indicators of changes in system states. The high accuracy in predicting vital signs without target conditional data shows that the model can extract interrelations among all features. The quick response of TDSTF is also crucial in the ICU setting. All these benefits help detect deteriorations in time, such as sudden tachycardia or developing hypotension, and thus are important for timely interventions in the ICU setting.

However, there is still room for improving TDSTF. One issue is that the limitations in the size of the Transformer input can hinder its effectiveness, as the memory and time consumption increases quadratically with its size \cite{ding2020cogltx}. Additionally, the data used in the study may introduce bias into the model as the eligible subjects were generally older and male (as shown in Table \ref{tab:patient}), which could limit its generalizability to other populations. These issues highlight the importance of considering both the technical limitations and the demographic characteristics of the data when developing and evaluating deep learning models for forecasting sparse time series.

We look forward to this work serving as a foundation for addressing practical clinical applications in the ICU. It is essential to notice that subtle variations in vital indicators are warning signals of clinical deterioration that could eventually result in adverse outcomes. To further improve treatment outcomes, we may extend the proposed TDSTF to include core body temperature, breathing rate, and oxygen saturation as additional forecasting outputs. Additionally, we may explore the application of the TDSTF in reducing false alarms, with the aim of alleviating the stress experienced by ICU caregivers.
\section{Conclusion}
The TDSTF model offers a solution to the limitations of current state-of-the-art probabilistic forecasting models. By incorporating a Transformer-based backbone and using a triplet method to avoid input noise and increase speed, TDSTF was able to outperform the main baseline model, CSDI, on both \textcolor{black}{SACRPS} and MSE by an average of $18.9\%$ and $34.3\%$, respectively, and more than $17$ times faster than CSDI in inference. The results of this study demonstrate the improved accuracy and efficiency of TDSTF for sparse time series forecasting, making it a more practical and valuable tool for forecasting vital signs in the ICU setting. Additionally, TDSTF's ability to capture temporal dependencies across all features further enhances its potential for real-world applications. Further studies of TDSTF’s performance should be performed in datasets with other patient characteristics. If further validated, performance in real-time scenarios would be indicated.\\[8pt]

\noindent \textbf{Conflict of Interest}\\
S.F.Q. is a consultant for Bryte Bed, Guidepoint Global, Cowen Services, Whispersom, DR Capital and Best Doctors and is a member of the Hypopnea Taskforce of the American Academy of Sleep Medicine. Other authors have nothing to disclose.\\[8pt]

\noindent \textbf{Data Availability Statement}\\
The MIMIC-III dataset used in this study is available at https://physionet.org/content/mimiciii/1.4/. In order to access the data, researchers must complete the application form.\\[8pt]

\noindent \textbf{Funding Statement}\\
This work was supported by grants from the National Heart, Lung, and Blood Institute (\#R21HL159661) and the National Science Foundation (\#2052528). Any opinions, findings, and conclusions or recommendations expressed in this material are those of the authors and do not necessarily reflect the views of the sponsors.

\bibliographystyle{unsrt}
\bibliography{references}

\begin{thebibliography}{10}

\bibitem{kenzaka2012importance}
Tsuneaki Kenzaka, Masanobu Okayama, Shigehiro Kuroki, Miho Fukui, Shinsuke
  Yahata, Hiroki Hayashi, Akihito Kitao, Daisuke Sugiyama, Eiji Kajii, and
  Masayoshi Hashimoto.
\newblock Importance of vital signs to the early diagnosis and severity of
  sepsis: association between vital signs and sequential organ failure
  assessment score in patients with sepsis.
\newblock {\em Internal Medicine}, 51(8):871--876, 2012.

\bibitem{yoon2019predicting}
Joo~Heung Yoon, Lidan Mu, Lujie Chen, Artur Dubrawski, Marilyn Hravnak,
  Michael~R Pinsky, and Gilles Clermont.
\newblock Predicting tachycardia as a surrogate for instability in the
  intensive care unit.
\newblock {\em Journal of Clinical Monitoring and Computing}, 33:973--985,
  2019.

\bibitem{subbe2001validation}
Christian~P Subbe, Michael Kruger, Peter Rutherford, and L~Gemmel.
\newblock Validation of a modified early warning score in medical admissions.
\newblock {\em Qjm}, 94(10):521--526, 2001.

\bibitem{sessler2019beyond}
Daniel~I Sessler and Bernd Saugel.
\newblock Beyond ‘failure to rescue’: the time has come for continuous ward
  monitoring.
\newblock {\em British journal of anaesthesia}, 122(3):304--306, 2019.

\bibitem{doig2011informing}
Alexa~K Doig, Frank~A Drews, and Maureen~R Keefe.
\newblock Informing the design of hemodynamic monitoring displays.
\newblock {\em CIN: Computers, Informatics, Nursing}, 29(12):706--713, 2011.

\bibitem{collins2012search}
Sarah~A Collins, Lena Mamykina, Desmond Jordan, Dan~M Stein, Alisabeth Shine,
  Paul Reyfman, and David Kaufman.
\newblock In search of common ground in handoff documentation in an intensive
  care unit.
\newblock {\em Journal of biomedical informatics}, 45(2):307--315, 2012.

\bibitem{kristinsson2022prediction}
{\AE}var~{\"O}rn Kristinsson, Ying Gu, S{\o}ren~M Rasmussen, Jesper
  M{\o}lgaard, Camilla Haahr-Raunkj{\ae}r, Christian~S Meyhoff, Eske~K Aasvang,
  and Helge~BD S{\o}rensen.
\newblock Prediction of serious outcomes based on continuous vital sign
  monitoring of high-risk patients.
\newblock {\em Computers in Biology and Medicine}, 147:105559, 2022.

\bibitem{ghassemi2015multivariate}
Marzyeh Ghassemi, Marco Pimentel, Tristan Naumann, Thomas Brennan, David
  Clifton, Peter Szolovits, and Mengling Feng.
\newblock A multivariate timeseries modeling approach to severity of illness
  assessment and forecasting in icu with sparse, heterogeneous clinical data.
\newblock In {\em Proceedings of the AAAI conference on artificial
  intelligence}, volume~29, 2015.

\bibitem{tipirneni2022self}
Sindhu Tipirneni and Chandan~K Reddy.
\newblock Self-supervised transformer for sparse and irregularly sampled
  multivariate clinical time-series.
\newblock {\em ACM Transactions on Knowledge Discovery from Data (TKDD)},
  16(6):1--17, 2022.

\bibitem{unanue2017recurrent}
Inigo~Jauregi Unanue, Ehsan~Zare Borzeshi, and Massimo Piccardi.
\newblock Recurrent neural networks with specialized word embeddings for
  health-domain named-entity recognition.
\newblock {\em Journal of biomedical informatics}, 76:102--109, 2017.

\bibitem{ij2018statistics}
H~Ij.
\newblock Statistics versus machine learning.
\newblock {\em Nat Methods}, 15(4):233, 2018.

\bibitem{liu2019early}
Shiyu Liu, Jia Yao, and Mehul Motani.
\newblock Early prediction of vital signs using generative boosting via lstm
  networks.
\newblock In {\em 2019 IEEE International Conference on Bioinformatics and
  Biomedicine (BIBM)}, pages 437--444. IEEE, 2019.

\bibitem{masum2019investigation}
Shamsul Masum, John~P Chiverton, Ying Liu, and Branislav Vuksanovic.
\newblock Investigation of machine learning techniques in forecasting of blood
  pressure time series data.
\newblock In {\em Artificial Intelligence XXXVI: 39th SGAI International
  Conference on Artificial Intelligence, AI 2019, Cambridge, UK, December
  17--19, 2019, Proceedings 39}, pages 269--282. Springer, 2019.

\bibitem{liu2021top}
Xiaoli Liu, Tongbo Liu, Zhengbo Zhang, Po-Chih Kuo, Haoran Xu, Zhicheng Yang,
  Ke~Lan, Peiyao Li, Zhenchao Ouyang, Yeuk~Lam Ng, et~al.
\newblock Top-net prediction model using bidirectional long short-term memory
  and medical-grade wearable multisensor system for tachycardia onset:
  algorithm development study.
\newblock {\em JMIR Medical Informatics}, 9(4):e18803, 2021.

\bibitem{phetrittikun2021temporal}
Ratchakit Phetrittikun, Kerdkiat Suvirat, Thanakron~Na Pattalung, Chanon
  Kongkamol, Thammasin Ingviya, and Sitthichok Chaichulee.
\newblock Temporal fusion transformer for forecasting vital sign trajectories
  in intensive care patients.
\newblock In {\em 2021 13th Biomedical Engineering International Conference
  (BMEiCON)}, pages 1--5. IEEE, 2021.

\bibitem{rasul2021autoregressive}
Kashif Rasul, Calvin Seward, Ingmar Schuster, and Roland Vollgraf.
\newblock Autoregressive denoising diffusion models for multivariate
  probabilistic time series forecasting.
\newblock In {\em International Conference on Machine Learning}, pages
  8857--8868. PMLR, 2021.

\bibitem{tashiro2021csdi}
Yusuke Tashiro, Jiaming Song, Yang Song, and Stefano Ermon.
\newblock Csdi: Conditional score-based diffusion models for probabilistic time
  series imputation.
\newblock {\em Advances in Neural Information Processing Systems},
  34:24804--24816, 2021.

\bibitem{sohl2015deep}
Jascha Sohl-Dickstein, Eric Weiss, Niru Maheswaranathan, and Surya Ganguli.
\newblock Deep unsupervised learning using nonequilibrium thermodynamics.
\newblock In {\em International Conference on Machine Learning}, pages
  2256--2265. PMLR, 2015.

\bibitem{wen2017multi}
Ruofeng Wen, Kari Torkkola, Balakrishnan Narayanaswamy, and Dhruv Madeka.
\newblock A multi-horizon quantile recurrent forecaster.
\newblock {\em arXiv preprint arXiv:1711.11053}, 2017.

\bibitem{salinas2020deepar}
David Salinas, Valentin Flunkert, Jan Gasthaus, and Tim Januschowski.
\newblock Deepar: Probabilistic forecasting with autoregressive recurrent
  networks.
\newblock {\em International Journal of Forecasting}, 36(3):1181--1191, 2020.

\bibitem{wang2019deep}
Yuyang Wang, Alex Smola, Danielle Maddix, Jan Gasthaus, Dean Foster, and Tim
  Januschowski.
\newblock Deep factors for forecasting.
\newblock In {\em International conference on machine learning}, pages
  6607--6617. PMLR, 2019.

\bibitem{jensen2022ensemble}
Vilde Jensen, Filippo~Maria Bianchi, and Stian~Normann Anfinsen.
\newblock Ensemble conformalized quantile regression for probabilistic time
  series forecasting.
\newblock {\em IEEE Transactions on Neural Networks and Learning Systems},
  2022.

\bibitem{yang2023diffusion}
Ling Yang, Zhilong Zhang, Yang Song, Shenda Hong, Runsheng Xu, Yue Zhao, Wentao
  Zhang, Bin Cui, and Ming-Hsuan Yang.
\newblock Diffusion models: A comprehensive survey of methods and applications.
\newblock {\em ACM Computing Surveys}, 56(4):1--39, 2023.

\bibitem{ho2020denoising}
Jonathan Ho, Ajay Jain, and Pieter Abbeel.
\newblock Denoising diffusion probabilistic models.
\newblock {\em Advances in Neural Information Processing Systems},
  33:6840--6851, 2020.

\bibitem{song2020denoising}
Jiaming Song, Chenlin Meng, and Stefano Ermon.
\newblock Denoising diffusion implicit models.
\newblock {\em arXiv preprint arXiv:2010.02502}, 2020.

\bibitem{austin2021structured}
Jacob Austin, Daniel~D Johnson, Jonathan Ho, Daniel Tarlow, and Rianne van~den
  Berg.
\newblock Structured denoising diffusion models in discrete state-spaces.
\newblock {\em Advances in Neural Information Processing Systems},
  34:17981--17993, 2021.

\bibitem{anand2022protein}
Namrata Anand and Tudor Achim.
\newblock Protein structure and sequence generation with equivariant denoising
  diffusion probabilistic models.
\newblock {\em arXiv preprint arXiv:2205.15019}, 2022.

\bibitem{blau2022threat}
Tsachi Blau, Roy Ganz, Bahjat Kawar, Alex Bronstein, and Michael Elad.
\newblock Threat model-agnostic adversarial defense using diffusion models.
\newblock {\em arXiv preprint arXiv:2207.08089}, 2022.

\bibitem{oord2016wavenet}
Aaron van~den Oord, Sander Dieleman, Heiga Zen, Karen Simonyan, Oriol Vinyals,
  Alex Graves, Nal Kalchbrenner, Andrew Senior, and Koray Kavukcuoglu.
\newblock Wavenet: A generative model for raw audio.
\newblock {\em arXiv preprint arXiv:1609.03499}, 2016.

\bibitem{kong2020diffwave}
Zhifeng Kong, Wei Ping, Jiaji Huang, Kexin Zhao, and Bryan Catanzaro.
\newblock Diffwave: A versatile diffusion model for audio synthesis.
\newblock {\em arXiv preprint arXiv:2009.09761}, 2020.

\bibitem{vaswani2017attention}
Ashish Vaswani, Noam Shazeer, Niki Parmar, Jakob Uszkoreit, Llion Jones,
  Aidan~N Gomez, {\L}ukasz Kaiser, and Illia Polosukhin.
\newblock Attention is all you need.
\newblock {\em Advances in neural information processing systems}, 30, 2017.

\bibitem{brown2020language}
Tom Brown, Benjamin Mann, Nick Ryder, Melanie Subbiah, Jared~D Kaplan, Prafulla
  Dhariwal, Arvind Neelakantan, Pranav Shyam, Girish Sastry, Amanda Askell,
  et~al.
\newblock Language models are few-shot learners.
\newblock {\em Advances in neural information processing systems},
  33:1877--1901, 2020.

\bibitem{zhang2018advances}
Cheng Zhang, Judith B{\"u}tepage, Hedvig Kjellstr{\"o}m, and Stephan Mandt.
\newblock Advances in variational inference.
\newblock {\em IEEE transactions on pattern analysis and machine intelligence},
  41(8):2008--2026, 2018.

\bibitem{feller2015theory}
William Feller.
\newblock On the theory of stochastic processes, with particular reference to
  applications.
\newblock In {\em Selected Papers I}, pages 769--798. Springer, 2015.

\bibitem{luo2022understanding}
Calvin Luo.
\newblock Understanding diffusion models: A unified perspective.
\newblock {\em arXiv preprint arXiv:2208.11970}, 2022.

\bibitem{devlin2018bert}
Jacob Devlin, Ming-Wei Chang, Kenton Lee, and Kristina Toutanova.
\newblock Bert: Pre-training of deep bidirectional transformers for language
  understanding.
\newblock {\em arXiv preprint arXiv:1810.04805}, 2018.

\bibitem{he2016deep}
Kaiming He, Xiangyu Zhang, Shaoqing Ren, and Jian Sun.
\newblock Deep residual learning for image recognition.
\newblock In {\em Proceedings of the IEEE conference on computer vision and
  pattern recognition}, pages 770--778, 2016.

\bibitem{johnson2016mimic}
Alistair Johnson, Tom Pollard, and Roger Mark.
\newblock Mimic-iii clinical database (version 1.4).
\newblock {\em PhysioNet}, 10(C2XW26):2, 2016.

\bibitem{lockwood2004vital}
Craig Lockwood, Tiffany Conroy-Hiller, and Tamara Page.
\newblock Vital signs.
\newblock {\em JBI Evidence Synthesis}, 2(6):1--38, 2004.

\bibitem{kingma2014adam}
Diederik~P Kingma and Jimmy Ba.
\newblock Adam: A method for stochastic optimization.
\newblock {\em arXiv preprint arXiv:1412.6980}, 2014.

\bibitem{paszke2019pytorch}
Adam Paszke, Sam Gross, Francisco Massa, Adam Lerer, James Bradbury, Gregory
  Chanan, Trevor Killeen, Zeming Lin, Natalia Gimelshein, Luca Antiga, et~al.
\newblock Pytorch: An imperative style, high-performance deep learning library.
\newblock {\em Advances in neural information processing systems}, 32, 2019.

\bibitem{de2015survey}
Wim De~Mulder, Steven Bethard, and Marie-Francine Moens.
\newblock A survey on the application of recurrent neural networks to
  statistical language modeling.
\newblock {\em Computer Speech \& Language}, 30(1):61--98, 2015.

\bibitem{ding2020cogltx}
Ming Ding, Chang Zhou, Hongxia Yang, and Jie Tang.
\newblock Cogltx: Applying bert to long texts.
\newblock {\em Advances in Neural Information Processing Systems},
  33:12792--12804, 2020.

\end{thebibliography}

\end{document}